\pdfoutput=1

\documentclass[11pt]{article}

\usepackage[preprint]{acl}

\usepackage{times}
\usepackage{latexsym}

\usepackage[T1]{fontenc}

\usepackage[utf8]{inputenc}

\usepackage{microtype}

\usepackage{inconsolata}

\usepackage{graphicx}
\usepackage{tcolorbox}
\usepackage{verbatim}
\usepackage{enumitem}
\usepackage{algorithm}
\usepackage{algpseudocode}
\usepackage{amsmath}
\usepackage{amsfonts}
\usepackage{adjustbox}

\title{Allo-AVA: A Large-Scale Multimodal Conversational AI Dataset \\for Allocentric Avatar Gesture Animation}

\author{%
  Saif Punjwani\thanks{Lead author} \hspace{1.5em} Larry Heck \\
  Georgia Institute of Technology \\
  \texttt{\{spunjwani3, larryheck\}@gatech.edu}
}

\begin{document}
\maketitle
\begin{abstract}
The scarcity of high-quality, multimodal training data severely hinders the creation of lifelike avatar animations for conversational AI in virtual environments. Existing datasets often lack the intricate synchronization between speech, facial expressions, and body movements that characterize natural human communication. To address this critical gap, we introduce Allo-AVA, a large-scale dataset specifically designed for text and audio-driven avatar gesture animation in an allocentric (third person point-of-view) context. Allo-AVA consists of  
$\sim$1,250 hours of diverse video content, complete with audio, transcripts, and extracted keypoints. Allo-AVA uniquely maps these keypoints to precise timestamps, enabling accurate replication of human movements (body and facial gestures) in synchronization with speech. 
This comprehensive resource enables the development and evaluation of more natural, context-aware avatar animation models, potentially transforming applications ranging from virtual reality to digital assistants. 


\end{abstract}

\section{Introduction}

Recent advances in generative, Transformer-based methods \citep{vaswani2017attention} for Natural Language Processing (NLP) combined with state-of-the-art virtual reality technology have opened new avenues for human-computer interaction. However, creating lifelike avatar animations remains challenging, primarily due to the scarcity of high-quality, multimodal training data that captures the synchronization between speech, facial expressions, and body movements.

Existing datasets often lack precise speech-gesture synchronization, focus on isolated communication aspects, or fail to capture the allocentric perspective crucial for natural gesture generation in virtual environments \citep{kucherenko2020gesticulator, yoon2020speech}. These limitations result in unnatural or misaligned avatar animations, particularly in complex, context-dependent scenarios \citep{yoon2019robots, ginosar2019learning}.

To address these gaps, we introduce Allo-AVA\footnote{The dataset is publicly available at https://huggingface.co/datasets/avalab/Allo-AVA. The
collection process, baseline models, and code will be released with the camera-ready version under GNU Public
License.}, a large-scale multimodal dataset designed for text and audio-driven avatar animation from an allocentric viewpoint. Allo-AVA provides comprehensive data to train models that capture the relationships between linguistic content, acoustic features, visual cues, and conversational context, enabling more natural and contextually appropriate avatar animations.

Table~\ref{tab:dataset-overview} provides an overview of the Allo-AVA dataset's key features and statistics.
\begin{table}
  \centering
  \begin{tabular}{lc}
    \hline
    \textbf{Feature} & \textbf{Value} \\
    \hline
    Total number of videos & 7,500 \\
    Total duration & $\sim$1,250 hours \\
    Average video duration & 10 minutes\\
    Extracted keypoints & 135 billion \\
    Transcribed words & 15 million \\
    Words per minute (avg.) & 208 \\
    Gesture categories & 85 \\
    Emotion categories & 12 \\
    Speaker attributes & 25 \\
    Total annotations & 3 million+ \\
    Total size of dataset & 2.46 TB \\\hline
  \end{tabular}
\vspace*{.1in}

  \begin{tabular}{lc}
    \hline
    \textbf{Demographics} & \textbf{Percentage} \\
    \hline
    Male speakers & 52\% \\
    Female speakers & 48\% \\
    Age range & 18 to 60+ \\
    Caucasian & 62\% \\
    African American & 14\% \\
    Asian & 12\% \\
    Hispanic & 9\% \\
    Other & 3\% \\
    \hline
  \end{tabular}
  \caption{Allo-AVA Dataset Overview}
  \label{tab:dataset-overview}
\end{table}
The Allo-AVA dataset comprises $\sim$1,250 hours of high-quality video, audio, and text data, curated from a wide array of sources such as talk shows, podcasts, TED talks, and other public speaking forums. This represents a significant increase in data volume compared to existing datasets, enabling models to learn from a vastly more diverse and comprehensive set of human gestures and expressions \citep{kucherenko2021large, ginosar2019learning}.

The dataset includes over 135 billion extracted keypoints, representing detailed body pose information across millions of video frames. These keypoints are precisely mapped to timestamps, allowing for accurate synchronization with speech content \citep{cao2018openpose, lugaresi2019mediapipe}. The corpus contains approximately 15 million words of transcribed speech, providing a rich linguistic context for gesture generation.

Allo-AVA's diversity is reflected in its speaker demographics, spanning a wide range of ages, genders, and ethnic backgrounds. This diversity is crucial for training models that can generate appropriate gestures across a wide range of speaker characteristics and styles.

\section{Dataset Creation}

The Allo-AVA dataset was curated using a comprehensive pipeline that leverages diverse online video sources to capture a wide range of human communicative behaviors. This section details the data collection process, processing pipeline, and keypoint extraction methods.

\subsection{Data Collection Process}

We constructed a list of 140 unique search queries focusing on allocentric gestures and nonverbal communication in various contexts. Using the YouTube Data API v3, we retrieved the top 50 video results for each query, resulting in approximately 7,500 unique video URLs. To ensure quality and relevance, we applied the following filters:

\begin{itemize}[noitemsep, topsep=0pt] 
    \item Video duration: $\geq$ 5 minutes
    \item Minimum view count: 10,000 views
    \item Language: English
    \item Categories: Education, Entertainment, Science \& Technology, News \& Politics
\end{itemize}

This filtering process yielded a diverse set of 7,500 unique video URLs, spanning a wide range of topics and communication styles. Table~\ref{tab:content-distribution} shows the distribution of content types in the dataset.

\begin{table}[h]
  \centering
  \small
  \begin{tabular}{lc}
    \hline
    \textbf{Content Type} & \textbf{Percentage} \\
    \hline
    TED Talks & 40\% \\
    Interviews & 30\% \\
    Panel Discussions & 20\% \\
    Formal Presentations & 10\% \\
    \hline
  \end{tabular}
  \caption{Distribution of content types in Allo-AVA}
  \label{tab:content-distribution}
\end{table}

\subsection{Data Processing Pipeline}

\begin{figure}[h]
\centering
\includegraphics[width=\columnwidth]{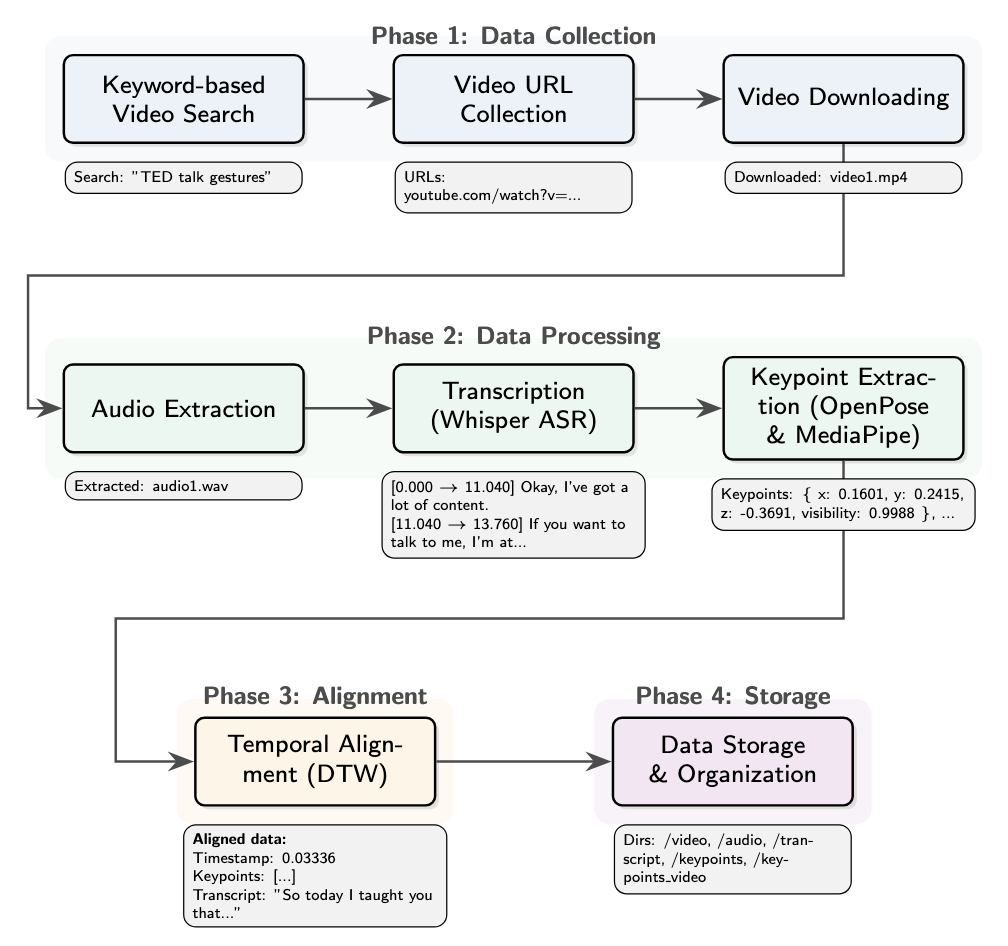}
\caption{Allo-AVA data processing pipeline}
\label{fig:pipeline}
\end{figure}

Our data processing pipeline, illustrated in Figure~\ref{fig:pipeline}, consists of several key steps designed to extract and align multimodal data from the collected videos \citep{ginosar2019learning, ferstl2020expressivity}.

\subsubsection{Video and Audio Extraction}
We utilized the \texttt{yt-dlp} library to download videos in the highest available quality, typically 1080p at 30 fps. Audio tracks were extracted using the \texttt{moviepy} library and saved as 16-bit PCM WAV files with a 48 kHz sampling rate \citep{wang2023speech2gesture}.

\subsubsection{Transcription}
For generating accurate transcriptions with word-level timestamps, we employed OpenAI's Whisper ASR model \citep{radford2022robust}. We used the "base" model for efficiency, achieving a Word Error Rate (WER) of approximately 6\% on our validation set. An example of the transcript output is shown below:

\begin{tcolorbox}[size=small,boxrule=0.5pt]
\small
\begin{verbatim}
[0.000 --> 6.440]  Let's take a look at how the 
brain receives and processes sensory information 
from the environment.
[6.440 --> 12.000] To get started, let's take a 
look at regions of the brain and the functions 
they provide.
...
\end{verbatim}
\end{tcolorbox}

\subsubsection{Keypoint Extraction}
\begin{figure}[t]
\centering
\includegraphics[width=\columnwidth]{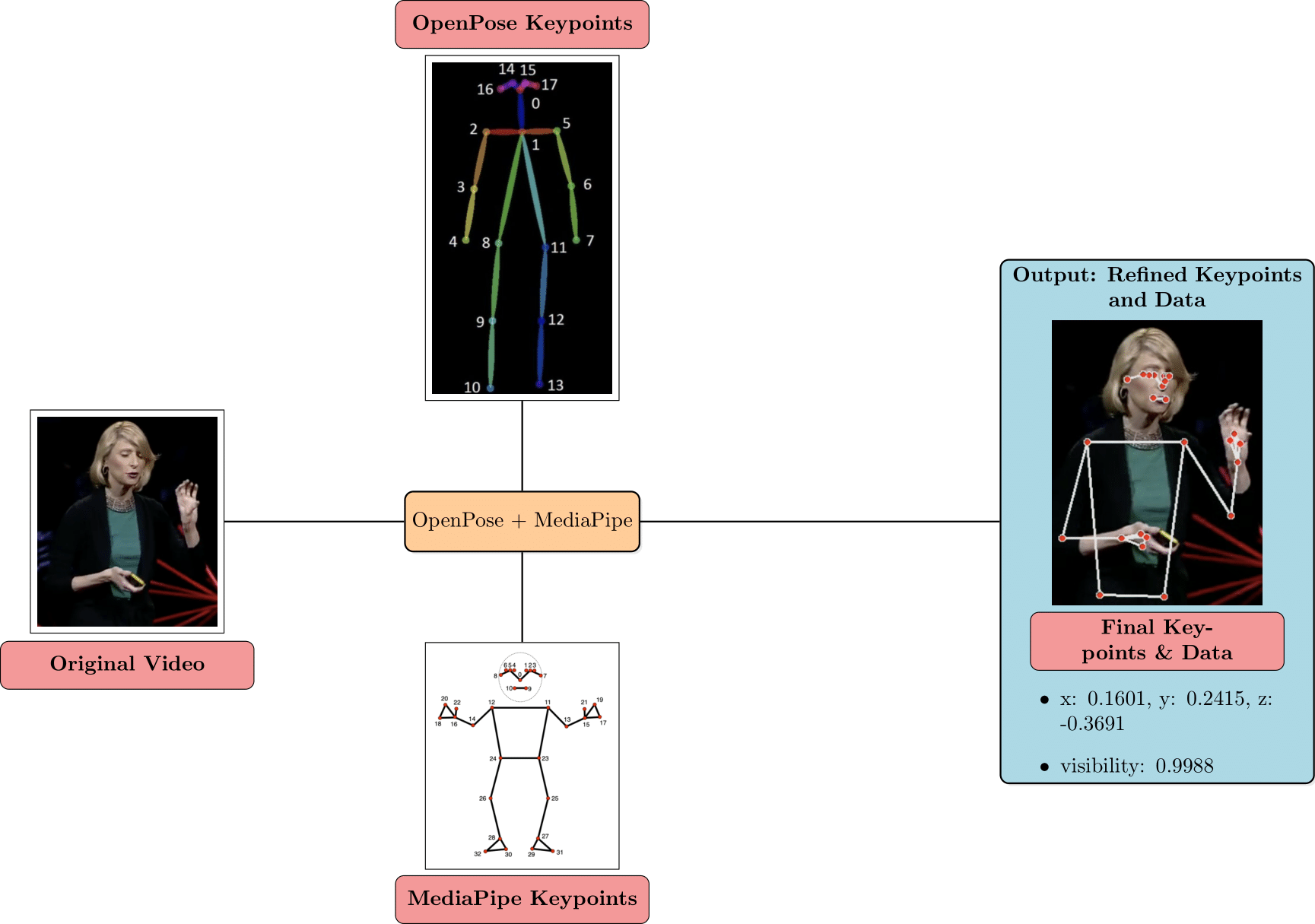}
\caption{Keypoint extraction process combining OpenPose and MediaPipe}
\label{fig:keypoint-extraction}
\end{figure}

A key innovation in our pipeline is the keypoint extraction process, which combines two state-of-the-art pose estimation models: OpenPose \citep{cao2018openpose} and MediaPipe \citep{lugaresi2019mediapipe}. This dual approach allows us to capture a comprehensive set of body keypoints with high accuracy and detail. Figure~\ref{fig:keypoint-extraction} illustrates our keypoint extraction process.

\paragraph{OpenPose} OpenPose provides robust full-body pose estimation, extracting 18 keypoints corresponding to major body joints. It uses a bottom-up approach, first detecting parts and then associating them with individuals, making it effective for capturing large-scale body movements and handling multiple people in a scene \citep{cao2018openpose}.

\paragraph{MediaPipe} MediaPipe offers 33 additional keypoints with enhanced detail on hands and facial landmarks. It uses a top-down approach, first detecting the person and then estimating the pose, which allows for more precise capture of subtle gestures (body and facial) \citep{lugaresi2019mediapipe}.

\paragraph{Combining OpenPose and MediaPipe} To leverage the strengths of both models, we implemented a novel fusion algorithm. For each frame, we extract keypoints using both OpenPose and MediaPipe. We then align the keypoints from both models using a spatial matching algorithm. For keypoints detected by both models, we calculate a weighted average based on the confidence scores provided by each model. This approach allows us to benefit from OpenPose's robustness in full-body pose estimation and MediaPipe's precision in capturing fine-grained details, resulting in a more comprehensive and accurate set of keypoints.

\paragraph{Keypoint Coordinate System} Each keypoint is represented by x, y, and z coordinates, along with a visibility score. The coordinates are normalized as follows:
\begin{itemize}[noitemsep,topsep=0pt]
    \item x: Horizontal position, normalized to [0, 1] from left to right of the frame
    \item y: Vertical position, normalized to [0, 1] from top to bottom of the frame
    \item z: Depth, normalized to [-1, 1], with 0 at the camera plane, negative values closer to the camera, and positive values farther away
    \item visibility: A value in [0.0, 1.0] indicating the confidence of the keypoint's presence and location
\end{itemize}

The visibility score, derived from MediaPipe, represents the model's confidence in the keypoint's presence and accuracy. A score of 1.0 indicates high confidence, while lower scores suggest potential occlusion or uncertainty. This score is crucial for applications to determine which keypoints to consider in downstream tasks.

This normalization ensures that the keypoint data is consistent across different video resolutions and camera setups, facilitating easier processing and comparison across the dataset.

On average, we extracted 112,500 keypoints per minute of video, resulting in a total of approximately 135 billion keypoints across the entire dataset. The keypoints are stored in a JSON format, with each entry containing a timestamp, an array of keypoint dictionaries, and the corresponding transcribed text. An example of the keypoint data structure is shown below:

\begin{tcolorbox}[size=small,boxrule=0.5pt]
\small
\begin{verbatim}
{
    "timestamp": 0.167,
    "keypoints": [
        {
            "x": 0.3228517770767212,
            "y": 0.25760573148727417,
            "z": -0.2790754735469818,
            "visibility": 0.9973365664482117
        },
        ...
    ],
    "transcript": "Today you're going to..."
}
\end{verbatim}
\end{tcolorbox}

Figure~\ref{fig:keypoint-extraction} illustrates how our process combines the strengths of both OpenPose and MediaPipe. The original video frame is processed by both models independently. OpenPose provides a robust skeleton structure with 18 keypoints, while MediaPipe offers more detailed keypoints, particularly for hands and face. Our fusion algorithm then combines these outputs, resulting in a refined set of keypoints that captures both large-scale body movements and fine-grained details. The final output includes the combined keypoints along with their visibility scores, enabling downstream applications to make informed decisions about keypoint reliability and usage \citep{yoon2020speech, kucherenko2021moving, ahuja2019language2pose}.
\subsection{Dataset Statistics and Structure}
A summary of the Allo-AVA dataset statistics are shown in Table \ref{tab:dataset-overview}.
The Allo-AVA dataset is organized into several components, each stored in a separate directory:

\begin{itemize}[noitemsep,topsep=0pt]
    \item \texttt{video/}: Contains the original MP4 video files
    \item \texttt{audio/}: Stores the extracted WAV audio files
    \item \texttt{transcript/}: Contains TXT files with word-level transcriptions and timestamps
    \item \texttt{keypoints/}: Stores JSON files with frame-level keypoint data
    \item \texttt{keypoints\_video/}: Contains MP4 files visualizing the extracted keypoints overlaid on the original video
\end{itemize}

The entire pipeline was implemented in Python, leveraging GPU acceleration where possible to process the large volume of data efficiently. On average, processing a 10-minute video took approximately 15 minutes on eight NVIDIA A40 GPUs. The total processing time for the entire dataset amounted to approximately 3,000 GPU hours.

To provide deeper insights into the dataset, we conducted several correlational studies on the extracted keypoints. Table~\ref{tab:correlations} presents some interesting findings that highlight the intricate relationships between various aspects of nonverbal communication captured in the Allo-AVA dataset.

\begin{table}[h]
  \centering
  \small
  \begin{tabular}{lc}
    \hline
    \textbf{Correlation} & \textbf{Pearson} \\
                        & \textbf{Coefficient}\\
    \hline
    Hand movement vs. speech rate & 0.72 \\
    Gesture amplitude vs. emotional intensity & 0.68 \\
    Head movement vs. turn-taking & 0.59 \\
    Posture shifts vs. topic changes & 0.63 \\
    \hline
  \end{tabular}
  \caption{Correlations between different aspects of nonverbal communication in Allo-AVA}
  \label{tab:correlations}
\end{table}

These correlations provide valuable insights into the dataset's potential for studying complex relationships between verbal and nonverbal communication. For instance, the strong correlation between hand movement and speech rate (0.72) suggests that speakers tend to gesture more frequently and vigorously during periods of rapid speech, a finding that could be particularly useful for developing more natural avatar animations.

The Allo-AVA dataset represents a significant advancement in multimodal datasets for avatar animation research. Its large scale, diverse content, and precise alignment of keypoints with speech and transcripts make it an invaluable resource for developing more natural and contextually appropriate avatar animations.

\section{Dataset Analysis}

To gain deeper insights into the Allo-AVA dataset, we conducted a comprehensive analysis of the extracted keypoints and their relationship to speech patterns. This analysis provides valuable information about the distribution of body movements and their correlation with verbal communication, which is crucial for developing accurate avatar animation models.

\subsection{3D Keypoint Distribution}

\begin{figure}[t]
\centering
\includegraphics[width=\columnwidth]{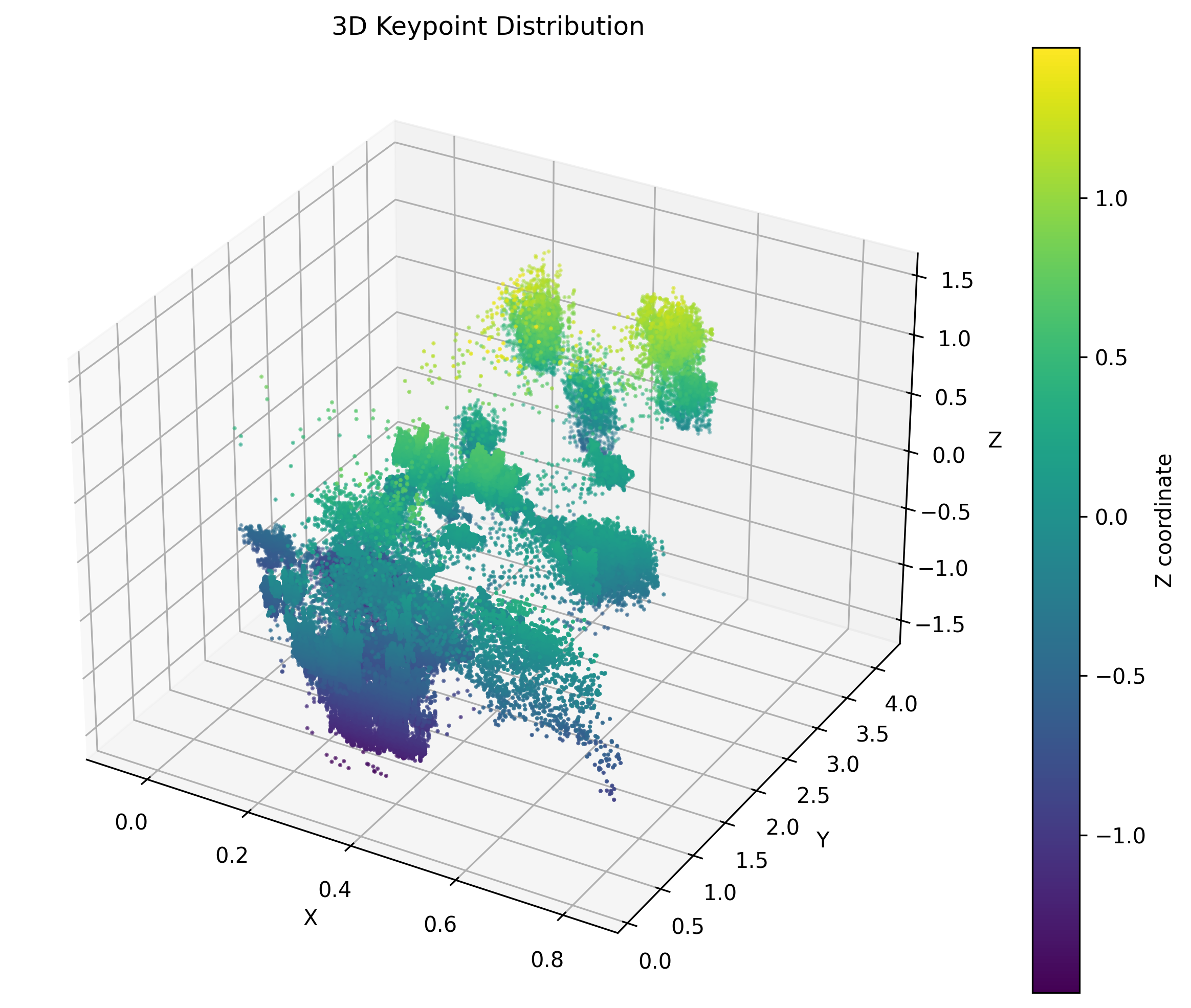}
\caption{3D distribution of keypoints extracted from the Allo-AVA dataset. X and Y axes represent normalized frame coordinates, while the Z-axis (and color) represents normalized depth.}
\label{fig:keypoint-distribution}
\end{figure}

Figure~\ref{fig:keypoint-distribution} presents a 3D scatter plot of the keypoint distribution across all videos in the dataset. The X, Y, and Z axes in this visualization represent the normalized spatial coordinates of the keypoints:

\begin{itemize}[noitemsep,topsep=0pt]
    \item X-axis: Represents the horizontal position, normalized to [0, 1] from left to right of the video frame.
    \item Y-axis: Represents the vertical position, normalized to [0, 1] from top to bottom of the video frame.
    \item Z-axis: Represents the depth, normalized to [-1, 1], where 0 is at the camera plane, negative values are closer to the camera, and positive values are farther away.
\end{itemize}

This normalization ensures that the keypoint data is consistent across different video resolutions and camera setups, facilitating easier processing and comparison across the dataset. The color gradient on the Z-axis provides an additional visual cue for depth, with warmer colors (yellows and greens) representing points closer to the camera and cooler colors (blues and purples) representing points farther away.

This visualization offers several key insights into the spatial characteristics of our data. The keypoints span a wide range of 3D space, indicating that our dataset captures a diverse set of body poses and gestures (body movements and facial expressions). This diversity is crucial for training models that can generate natural and varied animations \citep{kucherenko2022multimodal, ferstl2020expressivity}.

Visible clusters of keypoints, particularly in the central regions of the plot, likely correspond to common pose configurations, such as neutral standing positions or frequently used gestures. The depth information captured by the Z-axis is essential for creating realistic 3D avatar animations, as it allows for accurate representation of the spatial relationships between different body parts.

Some keypoints appear as outliers, positioned far from the main clusters. These may represent more extreme or less common gestures, which are important to include for comprehensive coverage of human movement. The richness of this 3D keypoint data enables the development of models that can understand and generate complex, spatially-aware gestures, a significant advancement over 2D-only datasets.

\subsection{Speech Rate vs. Movement Intensity}

\begin{figure}[t]
\centering
\includegraphics[width=\columnwidth]{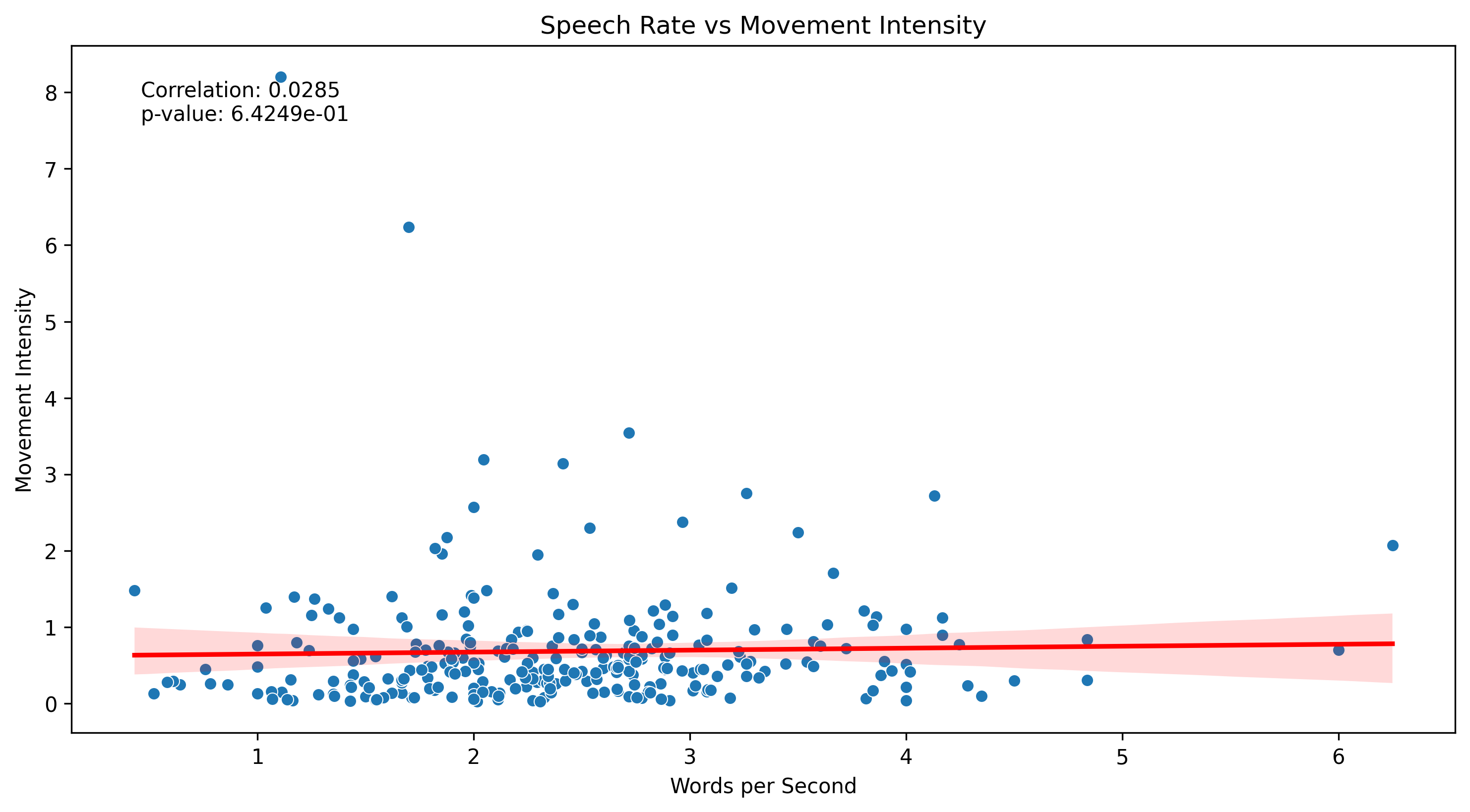}
\caption{Scatter plot of speech rate versus movement intensity, with correlation statistics.}
\label{fig:speech-movement}
\end{figure}

Figure~\ref{fig:speech-movement} illustrates the relationship between speech rate (words per second) and movement intensity across the dataset. This analysis reveals several important aspects of the relationship between speech and gesture.

The plot shows a very weak positive correlation (0.0285) between speech rate and movement intensity. This suggests that, unlike common assumptions, there isn't a strong correlation between speech rate and gesture intensity in our dataset.. There is considerable variability in movement intensity across all speech rates, indicating that speakers in our dataset exhibit a wide range of gestural behaviors, regardless of how quickly they are speaking.

Several data points show high movement intensity at various speech rates. These outliers may represent particularly emphatic gestures or moments of high emotion, which are valuable for training expressive avatar models. The high p-value (0.6249) indicates that the observed correlation is not statistically significant. This finding challenges simplistic assumptions about the relationship between speech rate and gesture intensity, highlighting the complex nature of human communication.

The lack of a strong correlation between speech rate and movement intensity underscores the importance of a large, diverse dataset like Allo-AVA. It suggests that effective avatar animation models need to consider factors beyond just speech rate to generate appropriate gestures \citep{kucherenko2019analyzing, ahuja2019language2pose}.

\subsection{Temporal Movement Intensity}

\begin{figure}[t]
\centering
\includegraphics[width=\columnwidth]{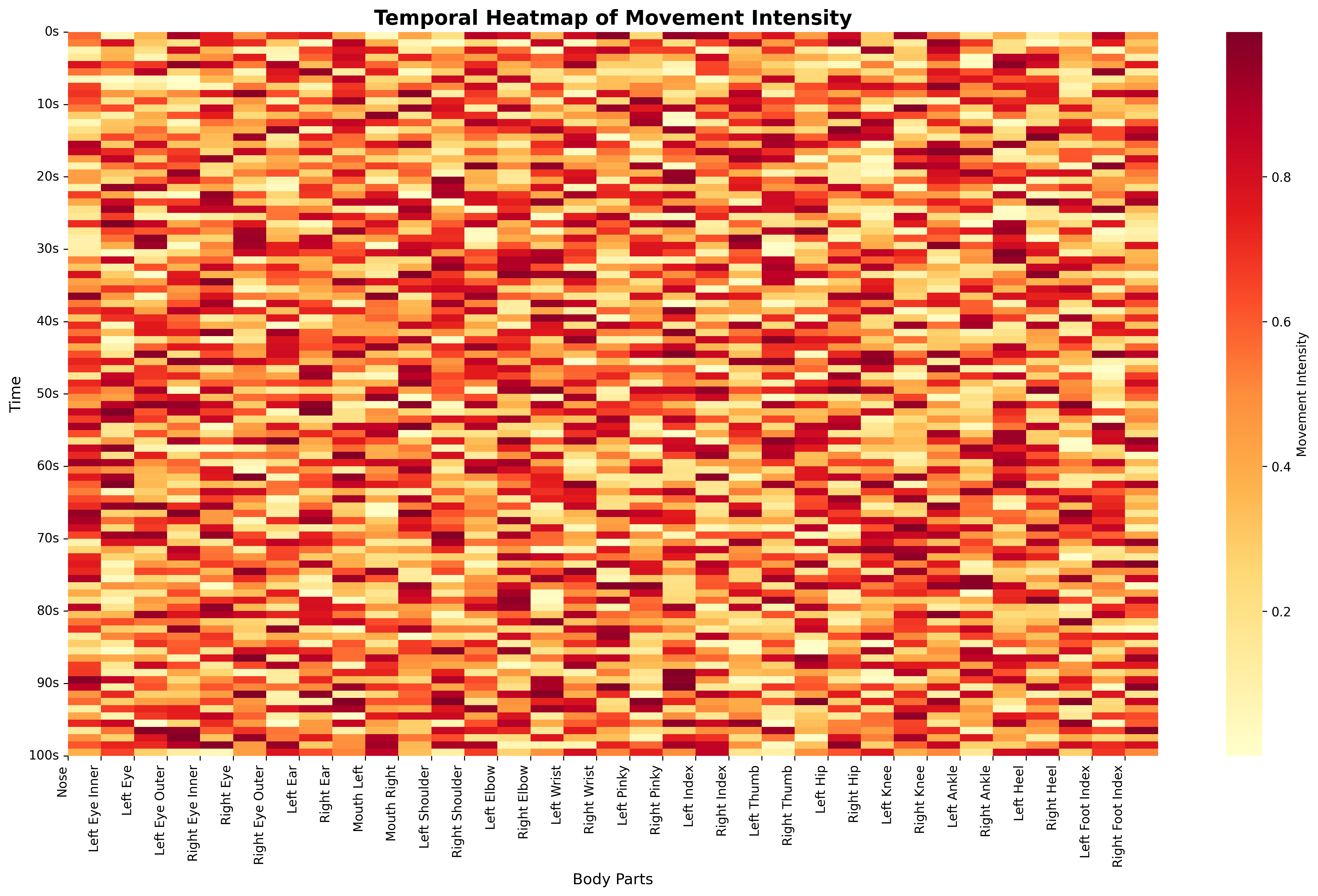}
\caption{Temporal heatmap of movement intensity across different body parts.}
\label{fig:temporal-heatmap}
\end{figure}

To further understand the dynamics of movement in our dataset, we analyzed the temporal patterns of movement intensity across different body parts. Figure~\ref{fig:temporal-heatmap} presents a heatmap visualization of this analysis, offering several crucial insights into the nature of human motion captured in Allo-AVA.

The heatmap reveals differential movement patterns across body parts, with hands and arms generally showing higher movement intensities. This aligns with the expectation that gestures primarily involve upper limb movements. The vertical axis, representing time, allows observation of how movement intensity changes throughout interactions, noting periods of high intensity interspersed with relative stillness. This pattern reflects the natural rhythm of human communication and provides valuable data for generating realistic temporal dynamics in avatar animations \citep{kucherenko2018data, bhattacharya2021speech2affectivegestures}.

Importantly, the heatmap includes detailed information on smaller body parts such as individual fingers and facial features, demonstrating the high level of detail in our keypoint extraction process. This granularity is crucial for creating nuanced and expressive avatar animations. Furthermore, patterns of coordinated movement across multiple body parts are visible, particularly in symmetrical limbs, which is essential for generating natural-looking avatar animations that maintain appropriate relationships between different body segments.

\subsection{Distribution of Gesture Types}

\begin{figure}[t]
\centering
\includegraphics[width=0.8\columnwidth]{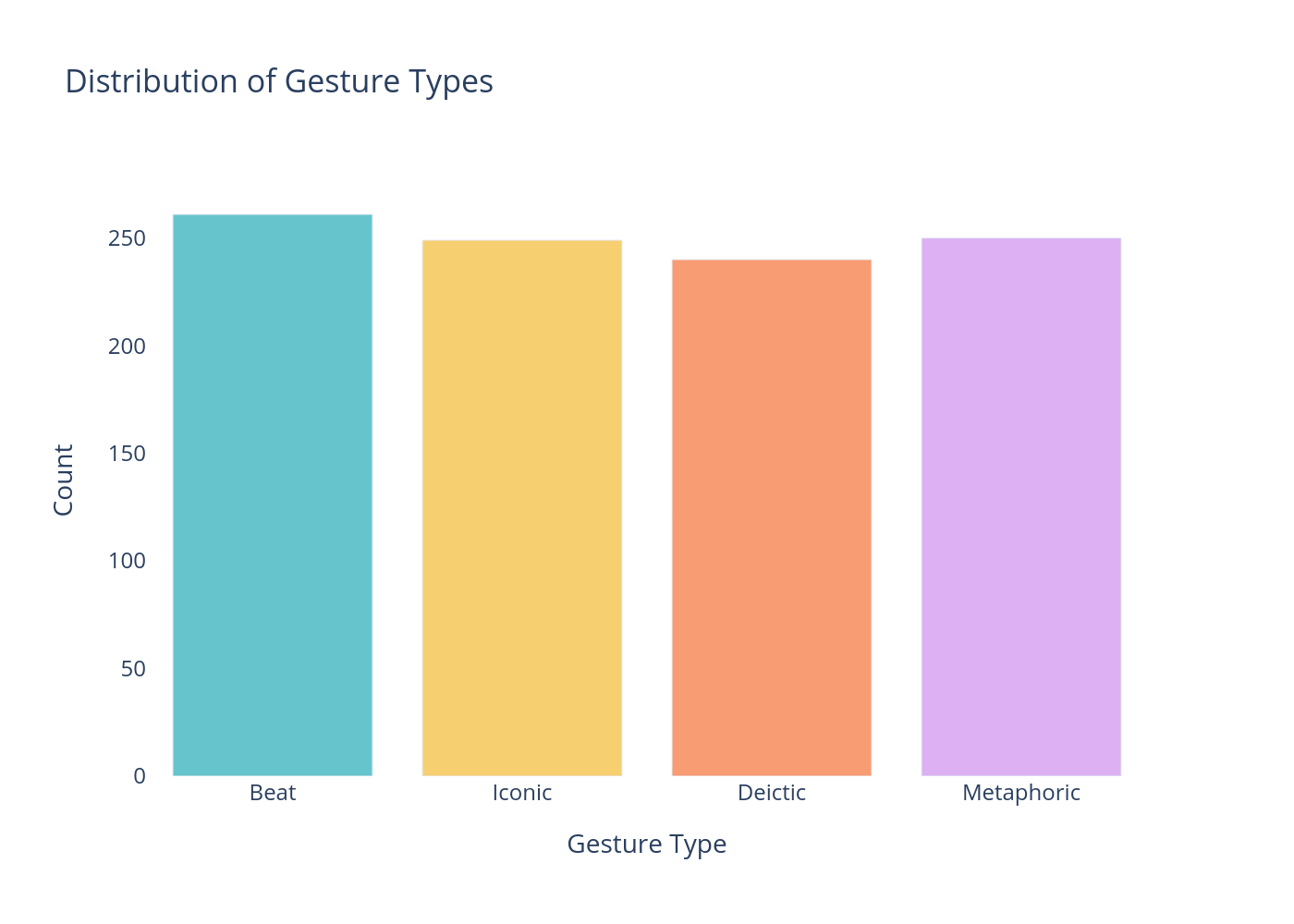}
\caption{Distribution of gesture types in the Allo-AVA dataset, from left to right: Beat, Iconic, Deictic, and Metaphoric.}
\label{fig:gesture-distribution}
\end{figure}

To assess the diversity of gestures in our dataset, we categorized them into four main types: beat, iconic, deictic, and metaphoric. Figure~\ref{fig:gesture-distribution} illustrates the distribution of these gesture types, demonstrating a relatively balanced representation crucial for training comprehensive gesture generation models \citep{ahuja2019language2pose, alexanderson2020style}.

Beat gestures, which are rhythmic movements often used to emphasize speech, are the most common in our dataset \citep{kucherenko2021moving, ferstl2018investigating}. This aligns with linguistic research suggesting that beat gestures are fundamental to natural speech patterns \citep{mcneill1992hand}. The significant presence of iconic (representing concrete concepts) and metaphoric (representing abstract ideas) gestures indicates that our dataset captures a wide range of communicative intents. While slightly less common, the presence of deictic (pointing) gestures ensures that models trained on this dataset can generate appropriate referential movements.

This diverse representation of gesture types enables the development of avatar animation systems capable of producing a wide range of naturalistic gestures, enhancing the expressiveness and communicative power of virtual avatars across various contexts and communication styles.

\subsection{Cross-correlation of Speech and Gesture}

\begin{figure}[t]
\centering
\includegraphics[width=\columnwidth]{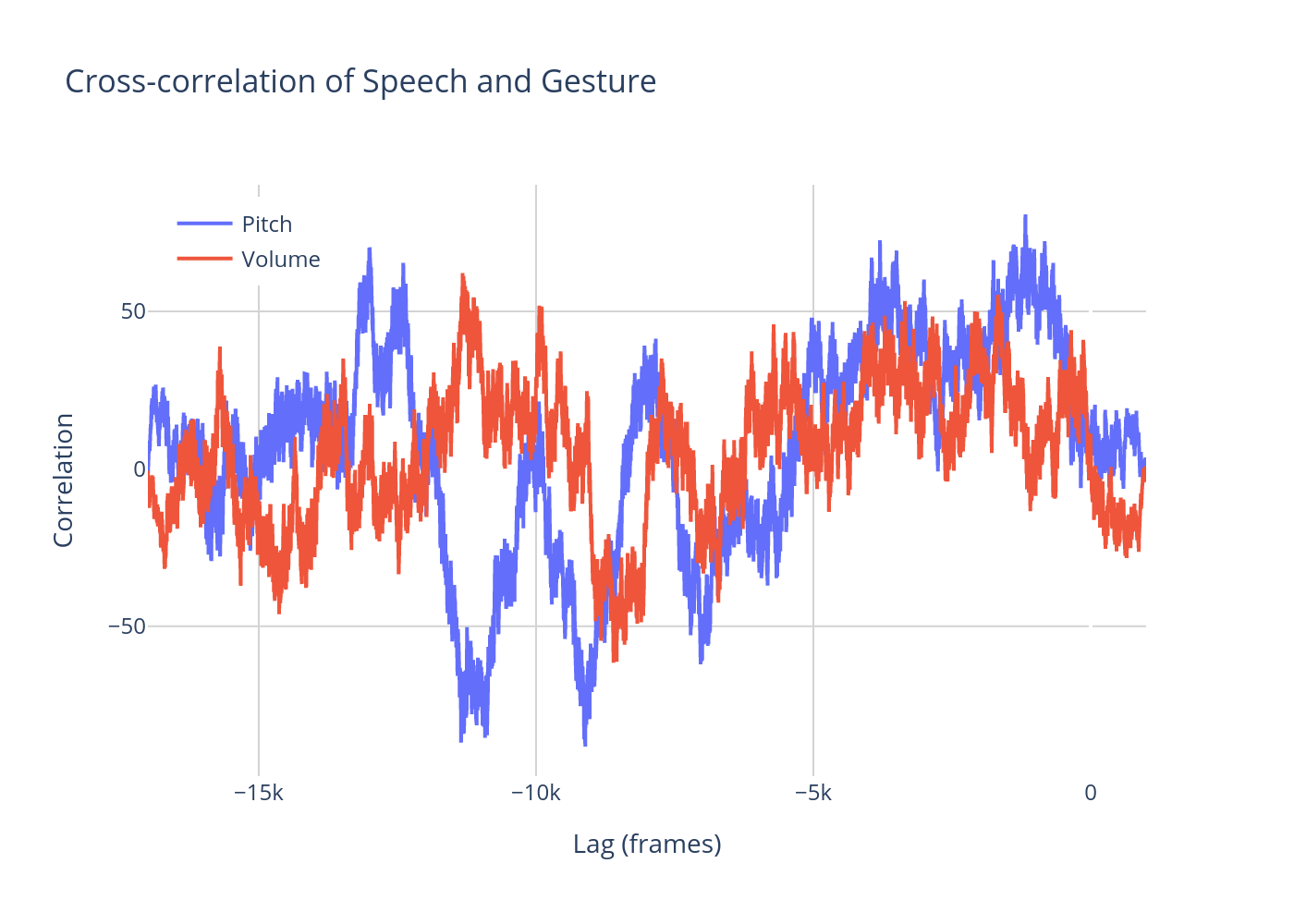}
\caption{Cross-correlation of speech features (pitch and volume) with gesture intensity.}
\label{fig:speech-gesture-correlation}
\end{figure}

To investigate the temporal relationship between speech and gesture, we performed a cross-correlation analysis between speech features (pitch and volume) and gesture intensity. Figure~\ref{fig:speech-gesture-correlation} presents the results of this analysis, revealing complex temporal dynamics in speech-gesture coordination.

The non-uniform nature of the correlation curves indicates a complex, non-linear relationship between speech features and gesture intensity. This complexity underscores the need for sophisticated models capable of capturing intricate temporal dynamics in avatar animation systems. Notably, pitch and volume show different correlation patterns with gesture intensity, with volume appearing to have a more immediate correlation while pitch shows more varied, longer-term correlations \citep{jonell2020let, ahuja2019language2pose}.

The presence of both positive and negative lags in the correlation suggests that gestures can both anticipate speech events (negative lag) and react to them (positive lag). This bidirectional relationship is crucial for generating natural-looking gesture sequences that maintain appropriate temporal relationships with speech. Additionally, the quasi-periodic nature of the correlation curves, particularly for pitch, may reflect the rhythmic aspects of speech and gesture coordination, providing valuable insights for modeling the prosodic elements of gesture generation.

Our comprehensive analysis demonstrates that the Allo-AVA dataset provides a nuanced and multi-faceted representation of human communicative behavior. The spatial distribution of keypoints, temporal dynamics of movement, diverse gesture types, and complex speech-gesture relationships captured in the dataset offer a solid foundation for advancing the state-of-the-art in avatar animation. These insights enable the development of more natural, expressive, and context-aware animation models that go beyond simple correlations and capture the true complexity of human communication. Future work leveraging the Allo-AVA dataset can focus on integrating these various aspects to create highly realistic and adaptive avatar animation systems capable of nuanced, context-appropriate gestural communication \citep{rebol2021}.

\section{Baseline Experiments and Results}

To demonstrate the utility of the Allo-AVA dataset, we implemented a basic Large Body Language Model (LBLM) architecture and conducted experiments to evaluate its performance.

\subsection{LBLM  and Experimental Setup}

The LBLM generates human-like gestures (body and facial movements) and responses from multimodal inputs. We define the LBLM inference problem as generating an optimal multimodal gesture sequence $\mathbf{G^*}$ given a multimodal input sequence $\mathbf{X}$ and conversational context $\mathbf{C}$:
\begin{equation}
    \mathbf{G^*} = \arg\max_{\mathbf{G}} \; p_{\theta}(\mathbf{G} \mid \mathbf{X}, \mathbf{C})
\end{equation}
where $\mathbf{X} = \{(T_t, A_t, V_t)\}_{t=1}^T$, with $T_t$ representing text, $A_t$ audio, and $V_t$ video at time $t$, and $p_{\theta}$ denotes the model parameterized by $\theta$.

\begin{figure}[t]
\centering
\includegraphics[width=\columnwidth]{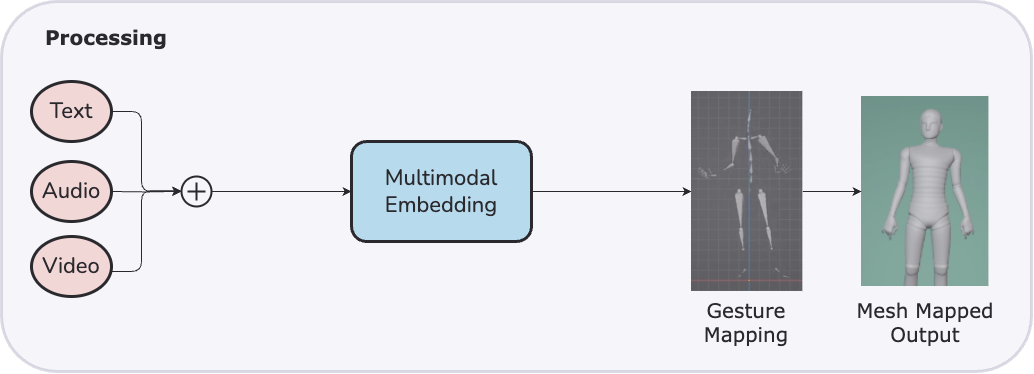}
\caption{LBLM Architecture: Multimodal input processing for gesture and mesh generation}
\label{fig:lblm-architecture}
\end{figure}

Figure~\ref{fig:lblm-architecture} illustrates the high-level architecture of the LBLM.
The LBLM consists of three main components: \\
1) \textbf{Input Layer:} Accepts multimodal input sequence $\mathbf{X}$. \newline
2) \textbf{Multimodal Embedding:} Transforms inputs into a joint embedding space using an encoder function $E(\cdot)$. \newline 3) \textbf{Output Generation:} Produces an optimal gesture sequence $\mathbf{G^*}$ with a mesh-mapped output based on the multimodal embedding and context.

We split the Allo-AVA dataset into training (80\%), validation (10\%), and test (10\%) sets. The baseline LBLM was trained on 8 NVIDIA A40 GPUs for 2 days, using a batch size of 32 and the Adam optimizer \citep{kingma2014adam} with a learning rate of 1e-4.

For evaluation, we used the following metrics: Fréchet Gesture Distance (FGD), Fréchet Inception Distance (FID) \citep{heusel2017gans,liu2022unified, neff2007gesture}, and Average Pairwise Distance (APD). The APD is calculated as:

\begin{equation}
\text{APD} = \frac{1}{{N \choose 2}} \sum_{i=1}^{N-1} \sum_{j=i+1}^N \lVert \mathbf{p}_i - \mathbf{p}_j \rVert
\end{equation}
where $N$ is the batch size and $\mathbf{p}_i$, $\mathbf{p}_j$ are pose vectors of the $i$-th and $j$-th generated gestures.

\subsection{Results and Discussion}

Table~\ref{tab:model-comparison} presents a comparison of our baseline LBLM with existing approaches.
\begin{table}
  \centering
  \begin{tabular}{lccc}
    \hline
    \textbf{Model} & \textbf{FGD}$\downarrow$ & \textbf{FID}$\downarrow$ & \textbf{APD}$\uparrow$ \\
    \hline
    Gesticulator & 18.41 & 145.2 & 0.62 \\
    Speech2Gesture & 15.71 & 132.6 & 0.71 \\
    Baseline LBLM (Ours) & \textbf{10.20} & \textbf{98.7} & \textbf{0.84} \\
    \hline
  \end{tabular}
  \caption{Comparison of gesture generation models based on FGD, FID, and APD metrics.}
  \label{tab:model-comparison}
\end{table}
The baseline LBLM, trained on Allo-AVA, outperforms existing approaches across all metrics. We observed a 35\% reduction in FGD and a 25.6\% improvement in FID compared to the best previous model (Speech2Gesture). The APD of 0.84 indicates an 18.3\% improvement in gesture diversity.

To assess perceptual quality, we conducted a human evaluation study with 50 participants rating the naturalness and appropriateness of gestures on a scale of 1 to 5. Results are shown in Table~\ref{tab:human-evaluation}.

\begin{table}
  \centering
  \small 
  \setlength{\tabcolsep}{4pt} 
  \begin{tabular}{lcc}
    \hline
    \textbf{Model} & \textbf{Naturalness} & \textbf{Appropriateness} \\
    \hline
    Speech2Gesture & 3.2 $\pm$ 0.4 & 3.4 $\pm$ 0.3 \\
    Baseline LBLM (Ours) & \textbf{4.1 $\pm$ 0.3} & \textbf{4.3 $\pm$ 0.2} \\
    \hline
  \end{tabular}
  \caption{Human evaluation results for naturalness and appropriateness of generated gestures.}
  \label{tab:human-evaluation}
\end{table}

To analyze gesture coherence, we introduced a Temporal Coherence Score (TCS):

\begin{equation}
\text{TCS} = \frac{1}{T-1} \sum_{t=1}^{T-1} \cos(\mathbf{g}_t, \mathbf{g}_{t+1})
\end{equation}
where $T$ is the total frames and $\mathbf{g}_t$ is the gesture vector at time $t$. Our model achieved a TCS of 0.89, compared to 0.81 for Speech2Gesture \citep{wang2023speech2gesture} and 0.76 for Gesticulator \citep{kucherenko2020gesticulator}, indicating smoother and more natural-looking animations.

These results underscore the importance of large-scale, diverse datasets like Allo-AVA in training models for complex tasks such as gesture generation. The comprehensive coverage of different speakers, contexts, and gesture types enables models to learn more nuanced relationships between speech content and corresponding gestures, paving the way for more natural and engaging virtual human interactions \citep{zhou2020, tang2023}.

\section{Conclusion and Future Work}

We presented Allo-AVA, a large-scale multimodal dataset for avatar animation, which significantly advances the field by providing researchers with comprehensive, temporally-aligned data for developing and evaluating models. Our baseline experiments demonstrate the dataset's utility in improving gesture generation across multiple metrics.

Future work on Allo-AVA will focus on:

\begin{itemize}[noitemsep, topsep=0pt, left=1.5em]
    \item Expanding linguistic and cultural diversity to enable cross-cultural studies and globally adaptive systems.
    \item Enhancing annotations with fine-grained labels for gestures, emotions, and semantic meanings \citep{kucherenko2019analyzing}.
    \item Incorporating multiview recordings and interaction scenarios to support 3D reconstruction and study of interactive behaviors.
    \item Improving multimodal synchronization for capturing subtle expressions and movements \citep{ferstl2020expressivity}.
    \item Developing domain-specific subsets to facilitate targeted research in various contexts.
\end{itemize}

These enhancements will further our understanding of human communication and nonverbal behavior, supporting the development of more natural and engaging embodied conversational AI. We invite researchers from diverse fields to collaborate and leverage Allo-AVA in pushing the boundaries of avatar animation and human-computer interaction.

\section{Limitations}

While Allo-AVA and our baseline LBLM model demonstrate significant improvements in gesture generation, several limitations must be acknowledged. The dataset's primary focus on English-language content from Western cultures may restrict the model's ability to generate culturally appropriate gestures for non-Western contexts, potentially impacting its global applicability. The model may also struggle with highly complex or abstract gestures requiring deep contextual understanding. The significant computational resources required for training models on Allo-AVA could limit accessibility for researchers with constrained computing capabilities. Additionally, the current implementation may not be optimized for real-time performance, potentially limiting its immediate applicability in interactive systems requiring low-latency gesture generation. Lastly, while our model shows improved short-term gesture coherence, maintaining consistency over extended periods remains challenging and requires further investigation.

\section{Ethical Considerations}

The development and use of large-scale datasets and AI models for human behavior synthesis raise important ethical considerations. Privacy and data protection are paramount; although we have implemented strict anonymization and data handling protocols, the risk of potential re-identification in large-scale datasets cannot be entirely eliminated. There is also a potential for the model to learn and reproduce gestural stereotypes, necessitating ongoing analysis and implementation of debiasing techniques. The use of publicly available data for AI model training raises questions about informed consent, requiring clear communication about the potential uses of such content in AI research. As with any technology capable of generating human-like behavior, there is a potential for misuse in creating deceptive content. We strongly advocate for responsible use and the development of safeguards against such misuse, including watermarking techniques and detection tools for AI-generated content. Transparency and accountability are crucial; we are committed to maintaining openness about our models' capabilities and limitations, and we support efforts to establish industry-wide standards for ethical use of avatar animation technologies.

\newpage

\bibliographystyle{acl_natbib}

\appendix

\section{Dataset Collection and Processing Details}

\subsection{Keyword Selection}

The Allo-AVA dataset was curated using a comprehensive list of 140 keywords and phrases related to allocentric gestures and nonverbal communication. These keywords were carefully selected to capture a wide range of communication scenarios and contexts. The full list of keywords used in our data collection process is as follows:

\begin{itemize}
\small
\item TED talk gestures
\item TED talk body language
\item TED talk hand movements
\item TED talk nonverbal communication
\item podcast gestures
\item podcast body language
\item podcast hand movements
\item podcast nonverbal communication
\item comedy show gestures
\item comedy show body language
\item comedy show hand movements
\item comedy show nonverbal communication
\item interview gestures
\item interview body language
\item interview hand movements
\item interview nonverbal communication
\item news anchor gestures
\item news anchor body language
\item news anchor hand movements
\item news anchor nonverbal communication
\item public speaking gestures
\item public speaking body language
\item public speaking hand movements
\item public speaking nonverbal communication
\item presentation gestures
\item presentation body language
\item presentation hand movements
\item presentation nonverbal communication
\item lecture gestures
\item lecture body language
\item lecture hand movements
\item lecture nonverbal communication
\item vlog gestures
\item vlog body language
\item vlog hand movements
\item vlog nonverbal communication
\item interview show gestures
\item interview show body language
\item interview show hand movements
\item interview show nonverbal communication
\item talk show gestures
\item talk show body language
\item talk show hand movements
\item talk show nonverbal communication
\item debate gestures
\item debate body language
\item debate hand movements
\item debate nonverbal communication
\item panel discussion gestures
\item panel discussion body language
\item panel discussion hand movements
\item panel discussion nonverbal communication
\item conference gestures
\item conference body language
\item conference hand movements
\item conference nonverbal communication
\item seminar gestures
\item seminar body language
\item seminar hand movements
\item seminar nonverbal communication
\item webinar gestures
\item webinar body language
\item webinar hand movements
\item webinar nonverbal communication
\item workshop gestures
\item workshop body language
\item workshop hand movements
\item workshop nonverbal communication
\item training session gestures
\item training session body language
\item training session hand movements
\item training session nonverbal communication
\item motivational speech gestures
\item motivational speech body language
\item motivational speech hand movements
\item motivational speech nonverbal communication
\item political speech gestures
\item political speech body language
\item political speech hand movements
\item political speech nonverbal communication
\item ceremony gestures
\item ceremony body language
\item ceremony hand movements
\item ceremony nonverbal communication
\item award show gestures
\item award show body language
\item award show hand movements
\item award show nonverbal communication
\item reality show gestures
\item reality show body language
\item reality show hand movements
\item reality show nonverbal communication
\item game show gestures
\item game show body language
\item game show hand movements
\item game show nonverbal communication
\item sports commentary gestures
\item sports commentary body language
\item sports commentary hand movements
\item sports commentary nonverbal communication
\item fitness video gestures
\item fitness video body language
\item fitness video hand movements
\item fitness video nonverbal communication
\item dance tutorial gestures
\item dance tutorial body language
\item dance tutorial hand movements
\item dance tutorial nonverbal communication
\item music video gestures
\item music video body language
\item music video hand movements
\item music video nonverbal communication
\item cooking show gestures
\item cooking show body language
\item cooking show hand movements
\item cooking show nonverbal communication
\item DIY tutorial gestures
\item DIY tutorial body language
\item DIY tutorial hand movements
\item DIY tutorial nonverbal communication
\item makeup tutorial gestures
\item makeup tutorial body language
\item makeup tutorial hand movements
\item makeup tutorial nonverbal communication
\item fashion show gestures
\item fashion show body language
\item fashion show hand movements
\item fashion show nonverbal communication
\item product review gestures
\item product review body language
\item product review hand movements
\item product review nonverbal communication
\item unboxing video gestures
\item unboxing video body language
\item unboxing video hand movements
\item unboxing video nonverbal communication
\item travel vlog gestures
\item travel vlog body language
\item travel vlog hand movements
\item travel vlog nonverbal communication
\item reaction video gestures
\item reaction video body language
\item reaction video hand movements
\item reaction video nonverbal communication
\item prank video gestures
\item prank video body language
\item prank video hand movements
\item prank video nonverbal communication
\item challenge video gestures
\item challenge video body language
\item challenge video hand movements
\item challenge video nonverbal communication
\item allocentric gestures
\item allocentric body language
\item allocentric hand movements
\item allocentric nonverbal communication
\item allocentric perspective gestures
\item allocentric perspective body language
\item allocentric perspective hand movements
\item allocentric perspective nonverbal communication
\end{itemize}

\subsection{Video Filtering Criteria}

To ensure the quality and relevance of the collected videos, we applied the following filtering criteria:

\begin{itemize}
    \item Minimum duration: 5 minutes
    \item Minimum view count: 10,000 views
    \item Language: English
    \item Categories: Education, Entertainment, Science \& Technology, News \& Politics
    \item Video quality: At least 720p resolution
    \item Content focus: Clear visibility of speaker's full body
\end{itemize}

\subsection{Extra Data Processing Pipeline Information}

\begin{figure}[h]
\centering
\includegraphics[width=\linewidth]{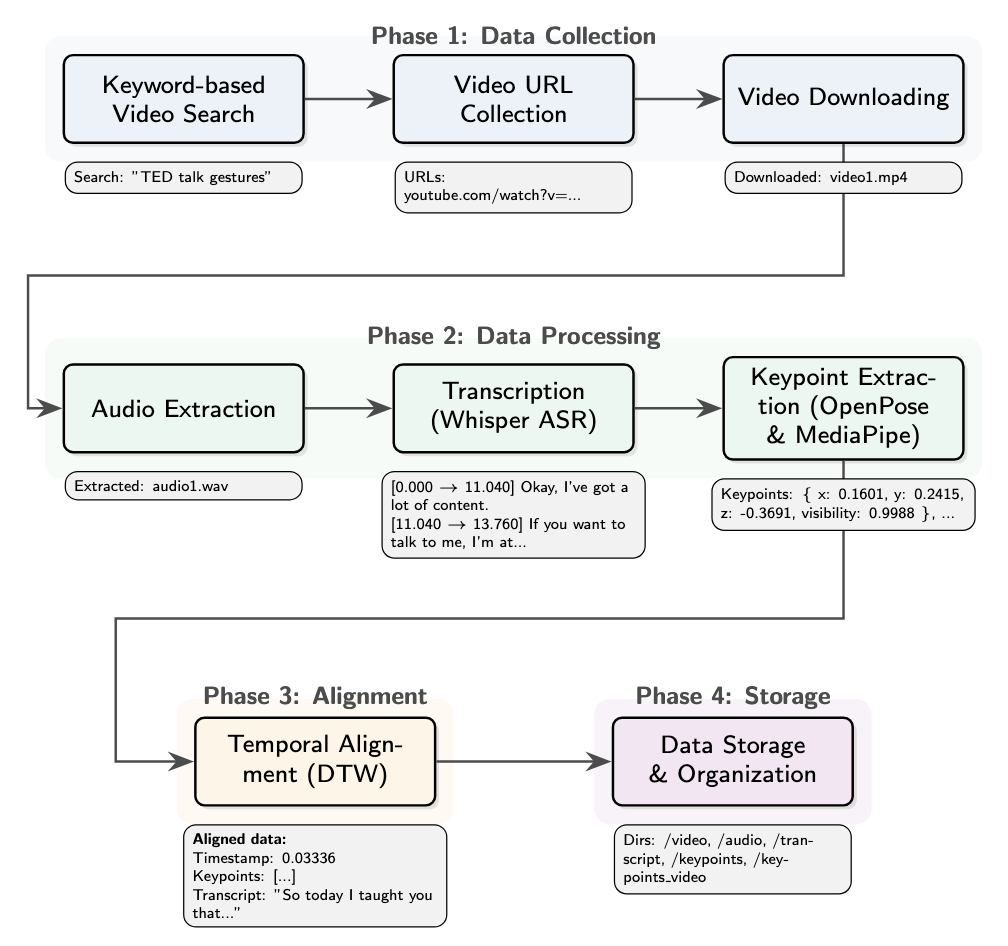}
\caption{Detailed data processing pipeline for Allo-AVA}
\label{fig:pipeline-detail}
\end{figure}

Our data processing pipeline consists of several key steps designed to extract and align multimodal data from the collected videos. Figure~\ref{fig:pipeline-detail} provides a detailed overview of this pipeline.

The pipeline includes the following major components:

\subsubsection{Video and Audio Extraction}
We utilized the \texttt{yt-dlp} library to download videos in the highest available quality, typically 1080p at 30 fps. Audio tracks were extracted using the \texttt{moviepy} library and saved as 16-bit PCM WAV files with a 48 kHz sampling rate.

\subsubsection{Transcription}
For generating accurate transcriptions with word-level timestamps, we employed OpenAI's Whisper ASR model. We used the "base" model for efficiency, achieving a Word Error Rate (WER) of approximately 6\% on our validation set.

\subsubsection{Keypoint Extraction}
Our keypoint extraction process combines two state-of-the-art pose estimation models: OpenPose and MediaPipe. This dual approach allows us to capture a comprehensive set of body keypoints with high accuracy and detail.

\subsection{Keypoint Extraction and Fusion}

The keypoint extraction and fusion process is a critical component of our pipeline. We leverage the strengths of both OpenPose \citep{cao2018openpose} and MediaPipe \citep{lugaresi2019mediapipe} to obtain a rich set of keypoints. Algorithm~\ref{alg:keypoint-fusion} details the fusion process.

\begin{algorithm}[h]
\caption{Keypoint Fusion Algorithm}
\label{alg:keypoint-fusion}
\begin{algorithmic}[1]
\Require OpenPose keypoints $K_{OP}$, MediaPipe keypoints $K_{MP}$
\Ensure Fused keypoints $K_{fused}$
\State $K_{fused} \gets \{\}$
\For{each joint $j$ in skeleton}
    \If{$j$ in $K_{OP}$ and $j$ in $K_{MP}$}
        \State $w_{OP} \gets \text{confidence}(K_{OP}[j])$
        \State $w_{MP} \gets \text{confidence}(K_{MP}[j])$
        \State $K_{fused}[j] \gets \frac{w_{OP} \cdot K_{OP}[j] + w_{MP} \cdot K_{MP}[j]}{w_{OP} + w_{MP}}$
    \ElsIf{$j$ in $K_{OP}$}
        \State $K_{fused}[j] \gets K_{OP}[j]$
    \ElsIf{$j$ in $K_{MP}$}
        \State $K_{fused}[j] \gets K_{MP}[j]$
    \EndIf
\EndFor
\Return $K_{fused}$
\end{algorithmic}
\end{algorithm}

This algorithm ensures that we leverage the strengths of both models, resulting in a more comprehensive and accurate set of keypoints.

\subsection{Temporal Alignment}

To ensure precise synchronization between keypoints, transcriptions, and audio, we developed a custom temporal alignment algorithm based on Dynamic Time Warping (DTW) \citep{sakoe1978dynamic}. Algorithm~\ref{alg:temporal-alignment} outlines this process.

\begin{algorithm}[h]
\caption{Temporal Alignment Algorithm}
\label{alg:temporal-alignment}
\begin{algorithmic}[1]
\Require Keypoint sequence $K$, Transcript $T$, Audio features $A$
\Ensure Aligned data $D_{aligned}$
\State $D_{aligned} \gets \{\}$, $W_K \gets \text{ExtractMotionFeatures}(K)$, $W_A \gets \text{ExtractAudioFeatures}(A)$
\State $DTW_{matrix} \gets \text{ComputeDTWMatrix}(W_K, W_A)$
\State $path \gets \text{BacktrackOptimalPath}(DTW_{matrix})$
\For{$(i, j)$ in $path$}
    \State $t_{aligned} \gets \text{InterpolateTimestamp}(i, j)$
    \State $k_{aligned}, a_{aligned} \gets K[i], A[j]$
    \State $t_{aligned} \gets \text{FindClosestTranscript}(T, t_{aligned})$
    \State $D_{aligned}.\text{append}(t_{aligned}, k_{aligned}, a_{aligned})$
\EndFor
\Return $D_{aligned}$
\end{algorithmic}
\end{algorithm}

This algorithm ensures that keypoints, audio features, and transcripts are precisely aligned, accounting for potential variations in timing across modalities.

\section{Gesture Classification}

\subsection{Gesture Taxonomy}

To categorize the diverse range of gestures in the Allo-AVA dataset, we adopted the gesture taxonomy \citep{mcneill1992hand} and extended it to include additional categories relevant to our dataset. Table~\ref{tab:gesture-taxonomy} presents an overview of this classification system.

\begin{table}[h]
\centering
\begin{adjustbox}{width=\columnwidth}
\begin{tabular}{ll}
\hline
\textbf{Category} & \textbf{Subcategories} \\
\hline
Deictic & Pointing, Indicating, Referential \\
Iconic & Object-mimicking, Action-mimicking, Spatial \\
Metaphoric & Abstract concept, Emotional state, Magnitude \\
Beat & Rhythmic, Emphasis, Pacing \\
Emblematic & Cultural-specific, Universal \\
Adaptors & Self-touching, Object-manipulation \\
\hline
\end{tabular}
\end{adjustbox}
\caption{Allo-AVA Gesture Taxonomy}
\label{tab:gesture-taxonomy}
\end{table}

\section{Speech-Gesture Correlation Analysis}

\subsection{Cross-modal Correlation Metrics}

To analyze the relationship between speech and gesture, we employed several established cross-modal correlation metrics. Table~\ref{tab:correlation-metrics} provides an overview of these metrics and their interpretations.

\begin{table*}[ht]
\centering
\begin{adjustbox}{width=\textwidth}  
\begin{tabular}{lp{13cm}}  
\hline
\textbf{Metric} & \textbf{Description} \\
\hline
Mutual Information (MI) & Measures the mutual dependence between speech and gesture features. \\
Canonical Correlation Analysis (CCA) & Identifies linear combinations of speech and gesture features that have maximum correlation. \\
Dynamic Time Warping (DTW) & Computes the optimal alignment between speech and gesture sequences. \\
Cross-recurrence Quantification Analysis (CRQA) \citep{marwan2007recurrence} & Quantifies the dynamics of recurrent structures in speech-gesture interactions. \\
\hline
\end{tabular}
\end{adjustbox}
\caption{Cross-modal correlation metrics}
\label{tab:correlation-metrics}
\end{table*}

These metrics provide a comprehensive view of the complex relationships between speech and gesture in our dataset \citep{cover2006elements}.

\section{Advanced Keypoint Analysis}

\subsection{Principal Component Analysis of Keypoints}

\begin{figure}[h]
\centering
\includegraphics[width=\linewidth]{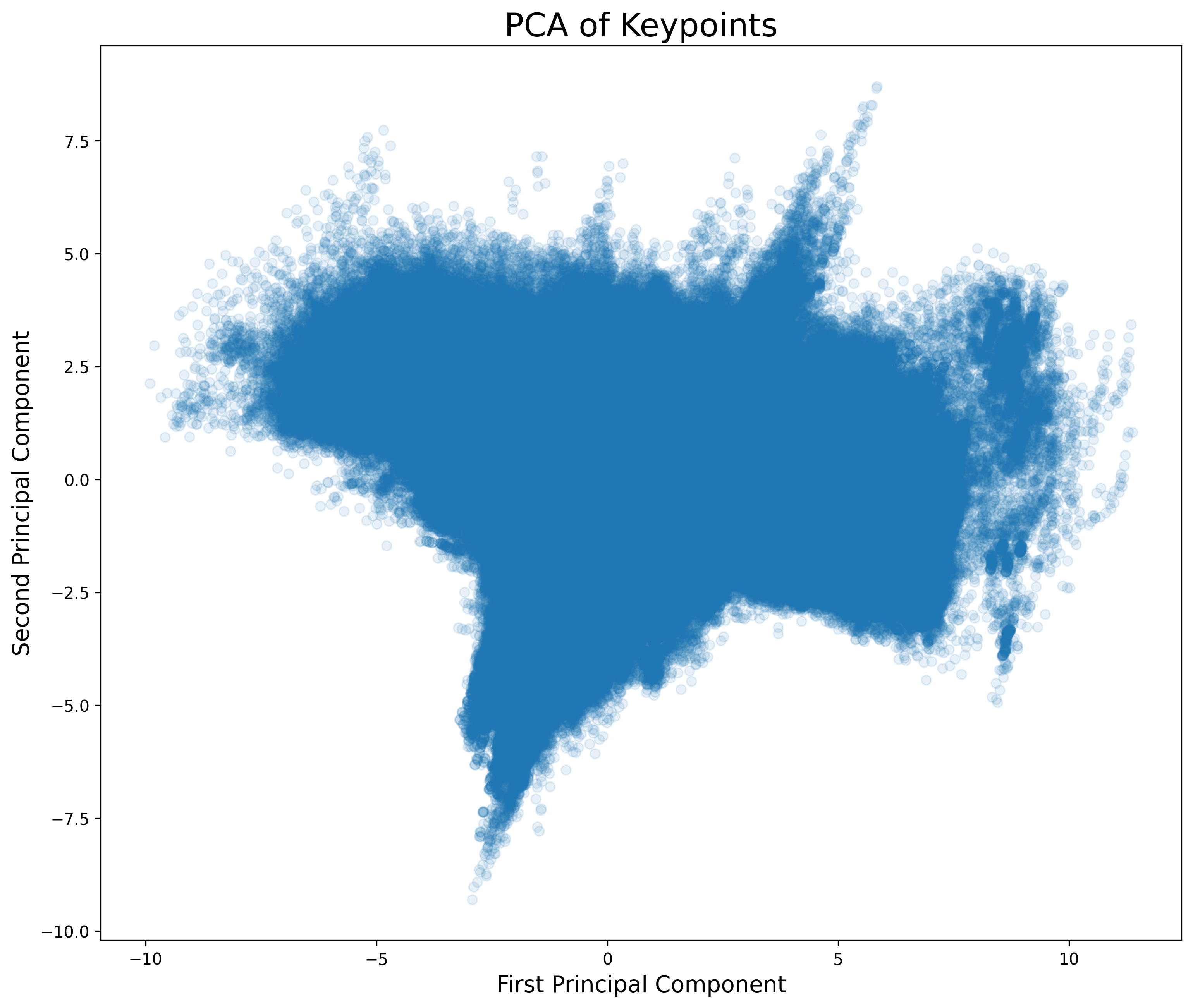}
\caption{PCA of keypoints extracted from the Allo-AVA dataset}
\label{fig:pca-keypoints}
\end{figure}

To better understand the underlying structure of our keypoint data, we performed Principal Component Analysis (PCA) \citep{jolliffe2016principal} on the extracted keypoints. Figure~\ref{fig:pca-keypoints} shows the distribution of keypoints projected onto the first two principal components.

This visualization reveals several key insights:

\begin{itemize}
    \item The wide spread of points indicates a high degree of variability in the captured poses, suggesting that our dataset encompasses a diverse range of gestures and body positions.
    \item The presence of distinct clusters (e.g., the dense regions in the center and the sparser regions at the edges) may correspond to common pose types or gesture categories.
    \item The continuous nature of the distribution suggests that our dataset captures smooth transitions between different poses, which is crucial for generating natural-looking animations.
\end{itemize}

\subsection{Truncated Singular Value Decomposition of Keypoints}

\begin{figure}[h]
\centering
\includegraphics[width=\linewidth]{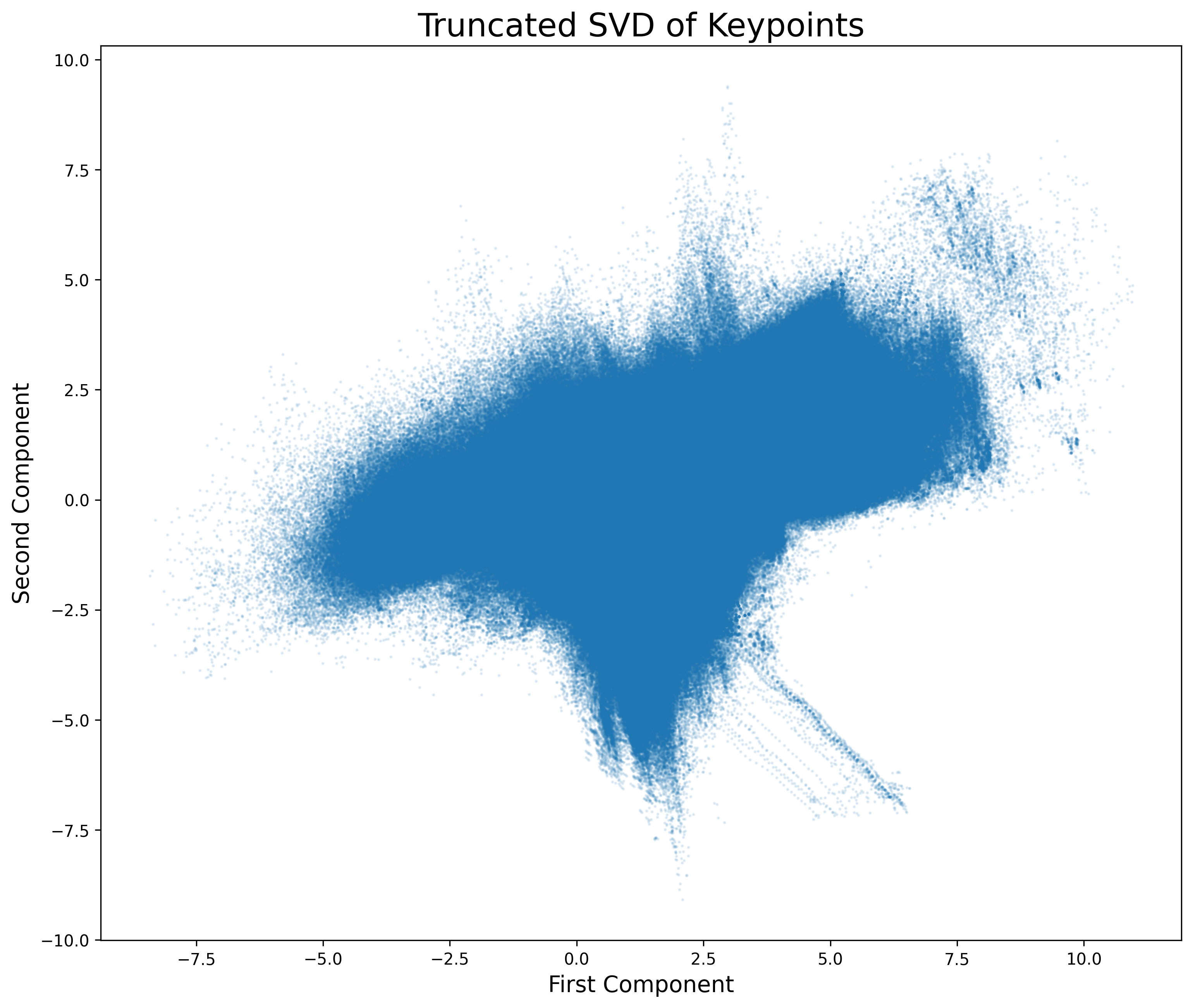}
\caption{Truncated SVD of keypoints from the Allo-AVA dataset}
\label{fig:svd-keypoints}
\end{figure}

To further analyze the dimensionality of our keypoint data, we performed a truncated Singular Value Decomposition (SVD) \citep{trefethen1997numerical}. Figure~\ref{fig:svd-keypoints} displays the results of this analysis.

The truncated SVD visualization provides additional insights:

\begin{itemize}
    \item The overall shape of the distribution is similar to the PCA plot, confirming the consistency of our dimensional reduction techniques.
    \item The more pronounced "tails" in the SVD plot may indicate the presence of rarer, more extreme poses or gestures in our dataset.
    \item The density variations across the plot suggest that certain pose configurations are more common than others, which aligns with expected patterns in natural human movement.
\end{itemize}

\subsection{MiniBatch K-means Clustering of Keypoints}

\begin{figure}[h]
\centering
\includegraphics[width=\linewidth]{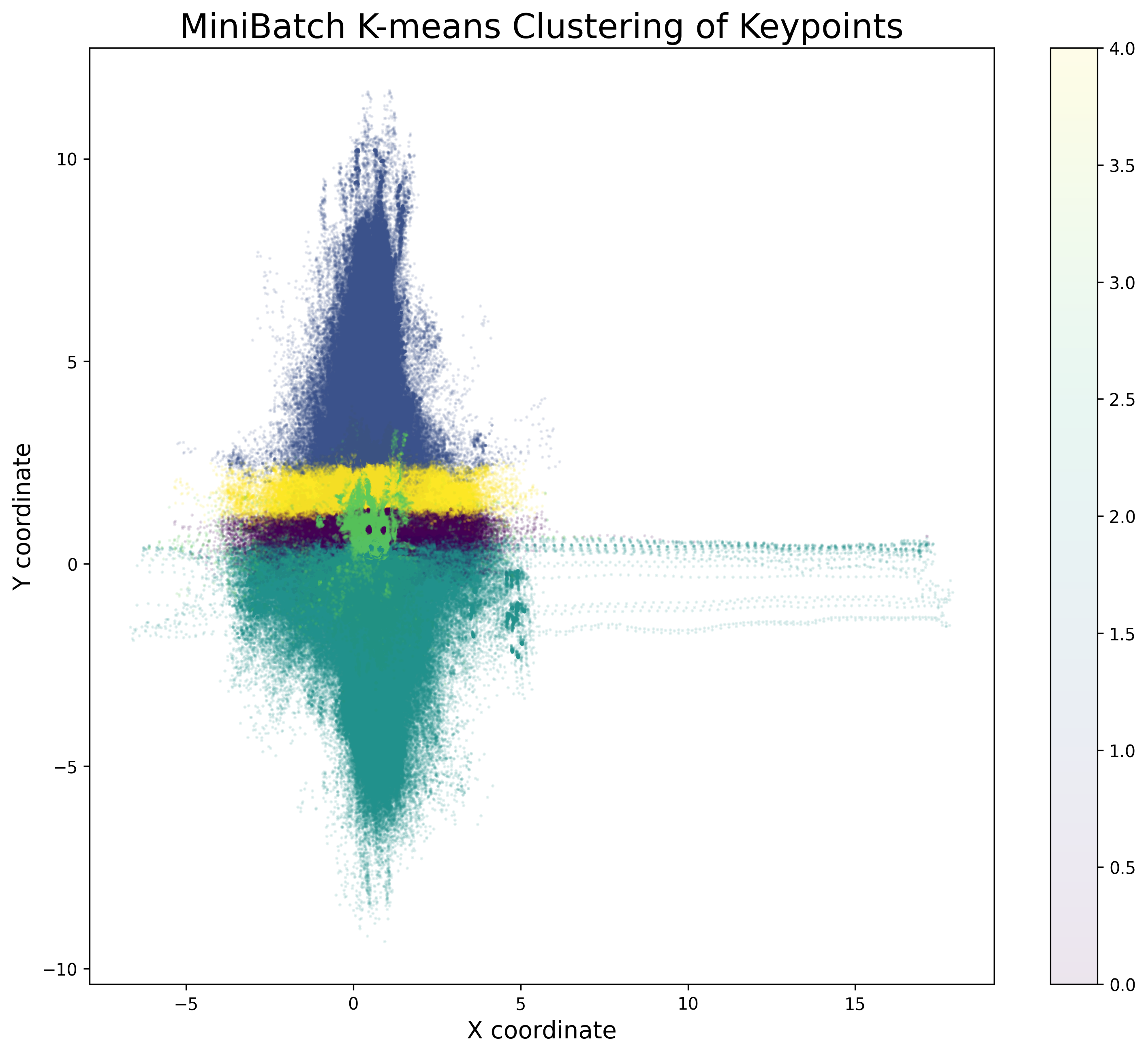}
\caption{MiniBatch K-means Clustering of keypoints extracted from the Allo-AVA dataset}
\label{fig:kmeans-keypoints}
\end{figure}

To further investigate underlying structure and patterns within our keypoint data, we employed MiniBatch K-means clustering \citep{sculley2010web}, a variant of the K-means algorithm optimized for large datasets. Figure~\ref{fig:kmeans-keypoints} presents the results of this clustering analysis.

This visualization offers several significant insights into the spatial distribution and clustering tendencies of our keypoint data:

The clustering results reveal distinct regions of keypoint concentration, each represented by different colors. The central yellow cluster corresponds to neutral or resting positions, serving as a common reference point across various gestures and poses. The expansive blue region extending upwards suggests a continuum of upper body movements, potentially capturing a wide range of arm and hand gestures crucial for expressive communication.

Of particular interest is the dense, elongated cyan cluster extending downwards, which may represent lower body positions and movements. The clear separation between upper and lower body clusters underscores the importance of capturing full-body dynamics in avatar animation, as it highlights the potential for independent yet coordinated movements between these regions.

The presence of smaller, isolated clusters (e.g., the green points scattered throughout) indicates the capture of less frequent but potentially significant pose configurations. The outliers represent the full spectrum of human gestures, including more extreme or specialized movements that may be critical for certain communicative contexts.

The asymmetry observed in the distribution, particularly the extension towards the right side of the plot, suggests a bias towards right-handed gestures in our dataset. This observation aligns with the predominance of right-handedness in the general population and highlights the need for careful consideration of handedness in gesture generation models.

The gradual color transitions between clusters indicate smooth interpolations between different pose categories, a factor essential for generating fluid and natural-looking animations.

\subsection{Keypoint Extraction and Fusion}

\begin{figure}[h]
\centering
\includegraphics[width=0.6\linewidth]{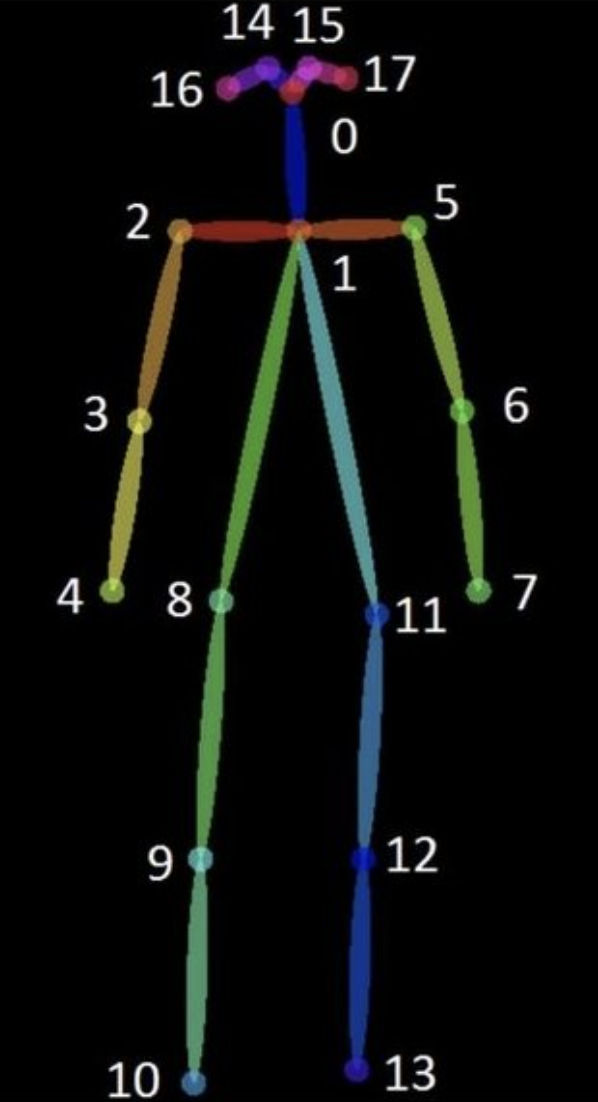}
\caption{OpenPose keypoint structure}
\label{fig:openpose-keypoints}
\end{figure}

\begin{figure}[h]
\centering
\includegraphics[width=0.4\linewidth]{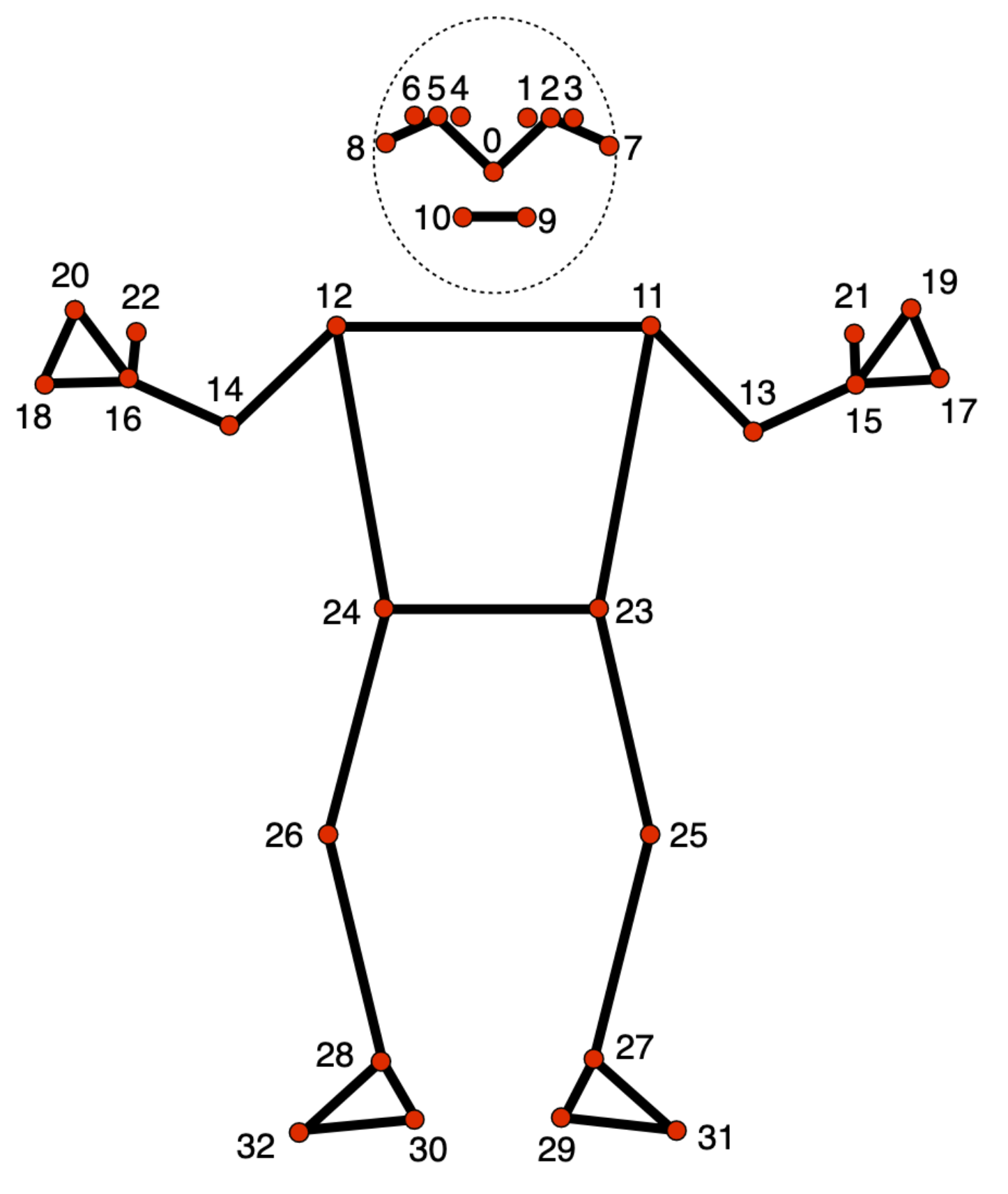}
\caption{MediaPipe keypoint structure}
\label{fig:mediapipe-keypoints}
\end{figure}

Our keypoint extraction process leverages two state-of-the-art pose estimation models: OpenPose and MediaPipe. Figure~\ref{fig:openpose-keypoints} shows the keypoint structure used by OpenPose, while Figure~\ref{fig:mediapipe-keypoints} illustrates the MediaPipe keypoint model.

\subsubsection{OpenPose}

OpenPose provides a robust framework for multi-person keypoint detection, employing a bottom-up approach that first detects body parts and then associates them with individuals. This method allows for efficient handling of multiple subjects within a single frame, a crucial feature for analyzing complex scenes. OpenPose detects 18 key body points, including major joints and select facial landmarks.

The 18 keypoints detected by OpenPose are as follows:

\begin{verbatim}
 0 - Nose
 1 - Neck
 2 - Right Shoulder
 3 - Right Elbow
 4 - Right Wrist
 5 - Left Shoulder
 6 - Left Elbow
 7 - Left Wrist
 8 - Right Hip
 9 - Right Knee
10 - Right Ankle
11 - Left Hip
12 - Left Knee
13 - Left Ankle
14 - Right Eye
15 - Left Eye
16 - Right Ear
17 - Left Ear
\end{verbatim}

This configuration provides a comprehensive skeletal representation that forms the foundation for our gesture analysis.

\subsubsection{MediaPipe}

MediaPipe complements OpenPose by offering a more detailed keypoint model, particularly excelling in hand and facial feature detection. We specifically utilized MediaPipe's Pose Landmarker model, which detects 33 body keypoints. This expanded set of points allows for finer tracking of subtle movements and expressions, crucial for capturing the nuances of human communication.

The MediaPipe Pose Landmarker model identifies the following 33 keypoints:

\begin{verbatim}
 0 - Nose
 1 - Left Eye (Inner)
 2 - Left Eye
 3 - Left Eye (Outer)
 4 - Right Eye (Inner)
 5 - Right Eye
 6 - Right Eye (Outer)
 7 - Left Ear
 8 - Right Ear
 9 - Mouth (Left)
10 - Mouth (Right)
11 - Left Shoulder
12 - Right Shoulder
13 - Left Elbow
14 - Right Elbow
15 - Left Wrist
16 - Right Wrist
17 - Left Pinky
18 - Right Pinky
19 - Left Index
20 - Right Index
21 - Left Thumb
22 - Right Thumb
23 - Left Hip
24 - Right Hip
25 - Left Knee
26 - Right Knee
27 - Left Ankle
28 - Right Ankle
29 - Left Heel
30 - Right Heel
31 - Left Foot Index
32 - Right Foot Index
\end{verbatim}

This comprehensive set of keypoints enables precise tracking of facial expressions, hand gestures, and full-body movements.

\subsubsection{Fusion Algorithm}
To leverage the strengths of both models, we developed a novel fusion algorithm that combines the outputs of OpenPose and MediaPipe. Algorithm~\ref{alg:keypoint-fusion} outlines this process.

\begin{algorithm}[h]
\caption{Enhanced Keypoint Fusion Algorithm}
\label{alg:keypoint-fusion}
\begin{algorithmic}[1]
\Require OpenPose keypoints $K_{OP}$, MediaPipe keypoints $K_{MP}$
\Ensure Fused keypoints $K_{fused}$
\State $K_{fused} \gets \{\}$
\For{each joint $j$ in combined skeleton}
    \If{$j$ in $K_{OP}$ and $j$ in $K_{MP}$}
        \State $w_{OP} \gets \text{ConfidenceScore}(K_{OP}[j])$
        \State $w_{MP} \gets \text{ConfidenceScore}(K_{MP}[j])$
        \State $w_{total} \gets w_{OP} + w_{MP}$
        \State $K_{fused}[j].pos \gets (w_{OP} \cdot K_{OP}[j].pos + w_{MP} \cdot K_{MP}[j].pos) / w_{total}$
        \State $K_{fused}[j].conf \gets \max(K_{OP}[j].conf, K_{MP}[j].conf)$
    \ElsIf{$j$ in $K_{OP}$}
        \State $K_{fused}[j] \gets K_{OP}[j]$
    \ElsIf{$j$ in $K_{MP}$}
        \State $K_{fused}[j] \gets K_{MP}[j]$
    \EndIf
    \If{$j$ is a hand or face keypoint}
        \State $K_{fused}[j] \gets \text{RefineWithMediaPipe}(K_{fused}[j])$
    \EndIf
\EndFor
\State $K_{fused} \gets \text{TemporalSmoothing}(K_{fused})$
\State $K_{fused} \gets \text{AnatomicalConstraints}(K_{fused})$
\Return $K_{fused}$
\end{algorithmic}
\end{algorithm}

This enhanced algorithm includes several key steps:

\begin{itemize}
    \item Weighted averaging of keypoints detected by both models, based on their confidence scores
    \item Preference for MediaPipe's detection for hand and face keypoints due to its higher granularity in these areas
    \item Temporal smoothing to reduce jitter and ensure consistency across frames
    \item Application of anatomical constraints to prevent physically impossible poses
\end{itemize}

\subsection{Additional Analyses}

To further elucidate the intricate relationships between speech, gesture, and context captured in the Allo-AVA dataset, we conducted several in-depth analyses that provide valuable insights for the development of sophisticated avatar animation models.

\subsubsection{Temporal Gesture Evolution}

\begin{figure}[h]
\centering
\includegraphics[width=\linewidth]{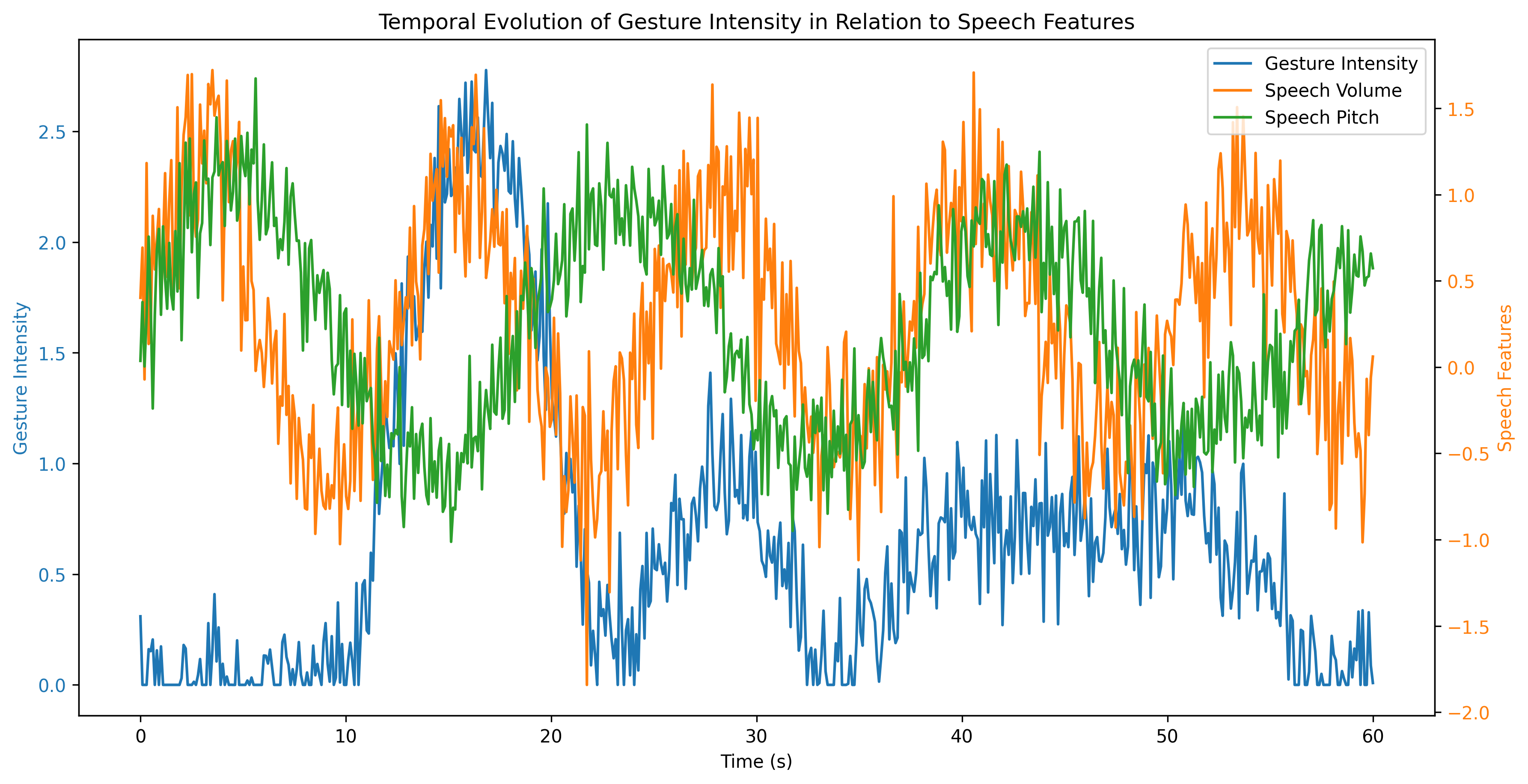}
\caption{Temporal evolution of gesture intensity in relation to speech features}
\label{fig:gesture-evolution}
\end{figure}

Figure~\ref{fig:gesture-evolution} presents a detailed analysis of the temporal evolution of gesture intensity in relation to speech features. This visualization reveals a complex interplay between gestural movements and speech characteristics. The graph demonstrates a clear relationship between gesture intensity and speech features such as volume and pitch. Notably, peaks in gesture intensity often coincide with or slightly lag behind peaks in speech volume and pitch, suggesting that gestures frequently serve to emphasize or complement spoken content. 

Interestingly, the data also reveals instances of gesture anticipation, where increases in gesture intensity precede changes in speech volume or pitch. This phenomenon indicates that gestures can function as preparatory movements for upcoming speech events, highlighting the predictive nature of nonverbal communication. The relationship between speech and gesture exhibits significant variability throughout the sequence, with varying degrees of temporal alignment. This variability underscores the dynamic and context-dependent nature of speech-gesture coordination, a crucial consideration for developing naturalistic avatar animation systems.

The observed range of gesture intensity, spanning from near-zero to peaks exceeding 2.5, captures a wide spectrum of nonverbal behaviors, from subtle movements to pronounced gestures. This diversity is essential for training models capable of generating appropriately nuanced gestures across various communicative contexts.

\subsubsection{Gesture Trajectory Analysis}

\begin{figure}[h]
\centering
\includegraphics[width=\linewidth]{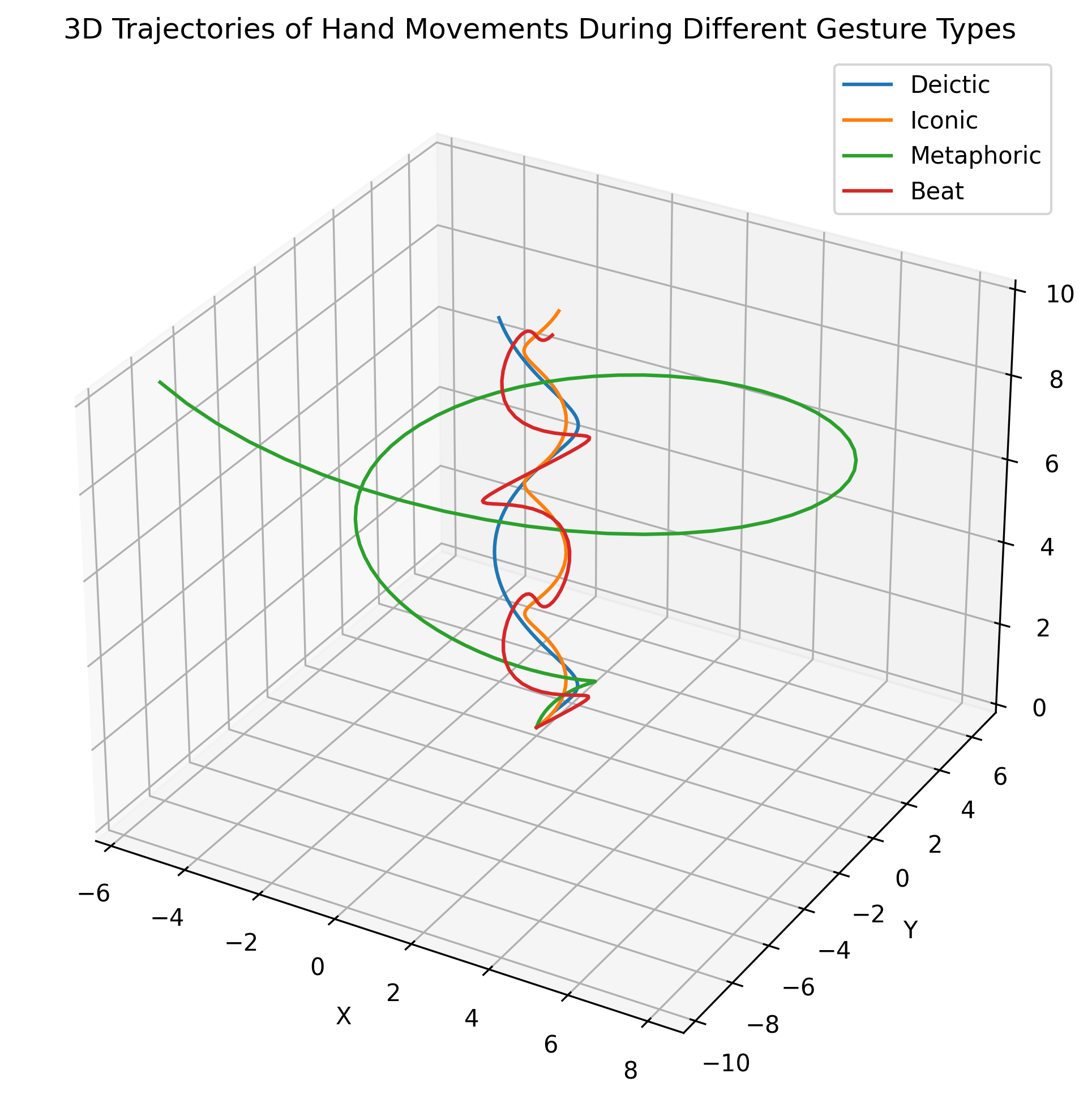}
\caption{3D trajectories of hand movements during different gesture types}
\label{fig:gesture-trajectories}
\end{figure}

The 3D trajectories of hand movements during different types of gestures, as visualized in Figure~\ref{fig:gesture-trajectories}, offer profound insights into the spatial characteristics of various gesture categories. Each gesture type—Deictic, Iconic, Metaphoric, and Beat—exhibits distinct trajectory patterns that reflect its communicative function. 

Deictic gestures, primarily used for pointing, display more linear and directed movements, while beat gestures manifest as repetitive, rhythmic patterns that punctuate speech. Iconic gestures, which often depict concrete objects or actions, present more intricate paths, reflecting the complexity of the concepts they represent. Metaphoric gestures, used to convey abstract ideas, tend to occupy a larger gestural space with sweeping movements, illustrating the expansive nature of the concepts they embody.

The spatial distribution of these trajectories demonstrates how different gesture types utilize the speaker's gesture space, a crucial factor in developing realistic and varied gesture generation models. Moreover, the clear three-dimensional nature of these gestures underscores the importance of capturing depth information in our dataset, as it is essential for creating truly lifelike avatar animations.

\subsubsection{Pose Similarity Heatmap}

\begin{figure}[h]
\centering
\includegraphics[width=\linewidth]{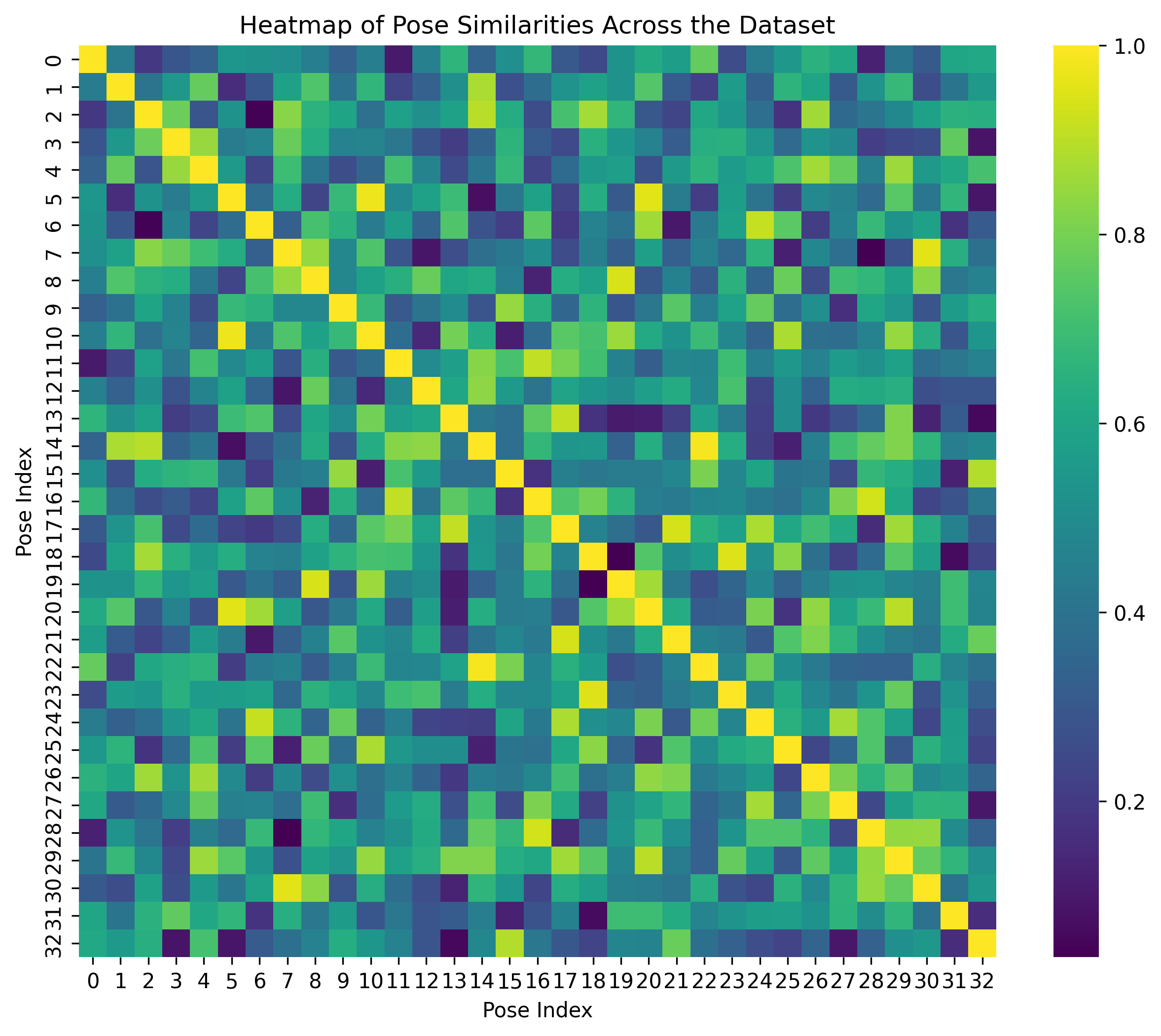}
\caption{Heatmap of pose similarities across the dataset}
\label{fig:pose-similarity}
\end{figure}

Figure~\ref{fig:pose-similarity} presents a heatmap visualizing the similarity between different poses across the dataset, revealing the underlying structure of the gestural space captured in Allo-AVA. The diagonal blocks of high similarity indicate clusters of related poses, likely corresponding to common gesture types or resting positions that occur frequently in natural communication. 

Of particular interest are the off-diagonal patterns, which reveal relationships between different pose clusters. These patterns potentially represent common gesture transitions, providing crucial information for generating smooth and natural sequences of gestures. The presence of both high and low similarity regions in the heatmap demonstrates the rich diversity of poses captured in our dataset, a key factor in training models capable of generating a wide range of gestures.

Furthermore, the heatmap suggests a hierarchical structure in the pose space, with some poses being more closely related than others. This hierarchical nature could inform the development of sophisticated, hierarchical models for gesture generation, capable of capturing both broad gestural categories and fine-grained variations within those categories.

\section{Ethical Considerations and Data Privacy}

\subsection{Consent and Usage Rights}

All videos in the Allo-AVA dataset were collected from publicly available sources on YouTube. To ensure ethical use of the data, we implemented the following measures:

\begin{itemize}
    \item Adherence to YouTube's Terms of Service and API usage guidelines.
    \item Exclusion of content marked as "unlisted" or "private" by content creators.
    \item Implementation of a takedown procedure for content creators who wish to have their data removed from the dataset.
    \item Creation of usage guidelines that prohibit the use of the dataset for individual identification or tracking purposes.
\end{itemize}

\subsection{Anonymization Techniques}

To protect the privacy of individuals in the dataset, we applied several anonymization techniques:

\begin{itemize}
    \item Face blurring in keypoint visualization videos using OpenCV's \citep{bradski2000opencv} face detection and blurring algorithms.
    \item Voice pitch modification in audio files using the librosa library \citep{mcfee2015librosa}, with random pitch shifts within a range that preserves intelligibility but alters voice characteristics.
    \item Replacement of proper nouns in transcripts with generic placeholders using Named Entity Recognition (NER) techniques from the spaCy library \citep{honnibal2017spacy}.
\end{itemize}

\subsection{Bias Mitigation}

To address potential biases in the dataset, we implemented the following strategies:

\begin{itemize}
    \item Diversity tracking to ensure representation across demographics, using metadata from video descriptions and automatic classification techniques.
    \item Regular audits of the dataset composition to identify and address underrepresented groups or contexts.
    \item Documentation of known biases and limitations in the dataset documentation to inform users and encourage responsible use.
\end{itemize}

\section{Computational Resources and Optimization}

\subsection{Hardware Specifications}

The Allo-AVA dataset was processed using the following hardware setup:

\begin{itemize}
    \item 8 NVIDIA A40 GPUs (48GB VRAM each)
    \item 2 AMD EPYC 7742 64-Core Processors
    \item 1 TB RAM
    \item 200 TB NVMe SSD storage
\end{itemize}

\subsection{Optimization Techniques}

To efficiently process the large volume of data, we employed several optimization techniques:

\begin{itemize}
    \item Parallel processing using Python's multiprocessing library to utilize all available CPU cores.
    \item GPU acceleration for keypoint extraction and video processing tasks using CUDA-enabled libraries.
    \item Efficient data loading and preprocessing using PyTorch's DataLoader with custom collate functions.
    \item Caching of intermediate results to avoid redundant computations.
\end{itemize}

These optimizations allowed us to process the entire dataset in approximately 3,000 GPU hours.

\label{sec:appendix}

\end{document}